\documentclass{article}

\usepackage[final, nonatbib]{neurips_2023}

\usepackage[utf8]{inputenc} %
\usepackage[T1]{fontenc}    %
\usepackage[USenglish]{babel}

\usepackage{microtype}
\usepackage{amsmath,amsfonts,amssymb,amsthm}
\usepackage{algorithmic}
\usepackage[linesnumbered,ruled]{algorithm2e}
\usepackage[toc,page]{appendix}
\usepackage{bm}
\usepackage{bbm}
\usepackage{booktabs} %
\usepackage{cancel}
\usepackage{caption} \usepackage{subcaption}
\usepackage{centernot}
\usepackage{csquotes}
\usepackage{enumitem}
\usepackage[mathscr]{eucal}
\usepackage{graphicx}
\usepackage{mathtools}
\usepackage{multirow}
\usepackage{nicefrac}       %
\usepackage{tabularx}
\usepackage{thmtools}
\usepackage{wrapfig}
\usepackage[table,xcdraw]{xcolor} \usepackage{tcolorbox}
\usepackage{tikz}
\usepackage{dsfont}

\usepackage[colorlinks,citecolor={green!80!black}]{hyperref}
\usepackage{url}

\usepackage[style=authoryear,sortcites,sorting=ynt,maxcitenames=2,maxbibnames=50,abbreviate=false,useprefix,uniquelist=minyear]{biblatex}
\let\citet\textcite  %
\let\citep\parencite
\renewbibmacro{in:}{}
\DeclareFieldFormat[unpublished,misc]{title}{\mkbibquote{#1}}  %
\DeclareNameAlias{sortname}{given-family}

\makeatletter %
\AtBeginDocument{\toggletrue{blx@useprefix}}
\AtBeginBibliography{\togglefalse{blx@useprefix}}
\makeatother

\usepackage[capitalize,noabbrev]{cleveref}

\makeatletter
\def\@captype{table}
\makeatother

\usepackage{thmtools}
\usepackage{thm-restate}

\newcommand{\bmf}[1]{\bm{\mathsf{#1}}}
\DeclareMathOperator*{\E}{\mathbb{E}}

\newcommand{\R}{\mathbb{R}}

\newcommand{\vd}{\bmf{d}}

\newcommand{\vw}{\bmf{w}}
\newcommand{\vx}{\bmf{x}}

\newcommand{\vz}{\bmf{z}}

\DeclareMathOperator*{\argmax}{argmax}
\newcommand{\red}[1]{\textcolor{red}{#1}}

\usepackage{physics}

\title{Bias Amplification in Language Model Evolution:\\An Iterated Learning Perspective}

\author{%
  Yi Ren\\
  UBC\\
  \texttt{renyi.joshua@gmail.com} \\
   \And
   Shangmin Guo \\
   University of Edinburgh\\
   \texttt{s.guo@ed.ac.uk} \\
   \And
   Linlu Qiu \\
   MIT \\
   \texttt{linluqiu@mit.edu} \\
   \And
   Bailin Wang \\
   MIT \\
   \texttt{bailin.wang28@gmail.com} \\
   \And
   Danica J. Sutherland \\
   UBC \& Amii \\
   \texttt{dsuth@cs.ubc.ca} \\
}

\begin{document}

\maketitle

\begin{abstract}
With the widespread adoption of Large Language Models (LLMs), the prevalence of iterative interactions among these models is anticipated to increase. Notably, recent advancements in multi-round on-policy self-improving methods allow LLMs to generate new examples for training subsequent models. At the same time, multi-agent LLM systems, involving automated interactions among agents, are also increasing in prominence. Thus, in both short and long terms, LLMs may actively engage in an evolutionary process. We draw parallels between the behavior of LLMs and the evolution of human culture, as the latter has been extensively studied by cognitive scientists for decades. Our approach involves leveraging Iterated Learning (IL), a Bayesian framework that elucidates how subtle biases are magnified during human cultural evolution, to explain some behaviors of LLMs. This paper outlines key characteristics of agents' behavior in the Bayesian-IL framework, including predictions that are supported by experimental verification with various LLMs. This theoretical framework could help to more effectively predict and guide the evolution of LLMs in desired directions.
The code for experiments is available at \url{https://github.com/Joshua-Ren/iICL}.
\end{abstract}

\section{Introduction}
\label{sec:intro}

Recent large language models (LLMs)
have shown remarkable instruction-following ability and an increasing number of applications;
it is thus reasonable to expect they are likely to become more widespread.
Moreover,
\emph{interactions} between LLMs (either multiple models, or different generations of the same model) may also become very commonplace in the near future.
In fact,
many recent works consider iterative on-policy self-data-augmentation solutions to break through the bottleneck of human-generated supervisions,
e.g., self-instruct \citep{wang2022self}, self-refine \citep{madaan2023selfrefine}, hypothesis refinement \citep{qiu2023phenomenal}, self-distill \citep{xu2023baize}, self-instruct \citep{wang2022self}, self-reward \citep{selfreward}, self-feedback \citep{xu2024perils}, RAFT \citep{dong2023raft}, ReST \citep{rest}, iterated DPO \citep{xiong2023gibbs}, OAIF \citep{guo2024direct}, SPIN \citep{chen2024self}, and many more.
Whether the model's knowledge is updated through in-weight or in-context mechanisms, 
these methods involve an LLM learning from a corpus (comprising data examples or evaluations) 
generated by another LLM (or itself), and subsequently transferring this acquired knowledge to others. 
Looking towards the long term, 
the future Internet may (for better or worse) contain substantial portions of LLM-generated text,
which will, in turn, 
be employed for training the subsequent generation of models. 
It thus seems important to begin studying what this process will mean for future models.

Although these self-improving methods demonstrate considerable improvements on various benchmarks,
a systematic understanding of why they work and what is their limitations is still missing.
Some analysis of knowledge distillation might bring insights \citep{mobahi2020self},
as learning from data generated by another model is a type of distillation.
But precisely analyzing the LLM's behavior on specific samples becomes increasingly difficult as it grows more complex.
Instead, a behavioral-level analysis might be fruitful,
akin to how the Bayesian framework can aid in comprehending the human cognitive system \citep{griffiths2023bayes}.
By conceptualizing the LLM as an intelligent agent,
we can draw parallels between its behaviors and the cultural evolution observed in humans.
Iterated learning (IL),
a framework proposed to study the evolution of knowledge and beliefs through a chain of learning among Bayesian agents \citep{kirby2007innateness}, 
stands out as a promising candidate for achieving our goals.

In this paper, we start by introducing the Bayesian-IL framework, 
demonstrating that agents engaged in such a process gradually amplify bias in their priors.
This amplification process can be steered by introducing an interaction phase that ``filters'' or ``re-ranks'' the messages generated by the agents.
Next,
we theoretically justify that the in-context behavior of LLMs can be approximated by a Bayesian update, 
establishing a crucial link to the LLM system.
To validate our claims, 
we conduct numerous experiments across different settings. 
Depending on the beneficial or detrimental nature of the bias, 
we propose various strategies to guide the evolution of LLM.
The key contributions of this work are:
1) establishing the first Bayesian analysis of the full interactive learning process (including an interaction phase);
2) applying this framework to LLM agents and describing their evolution theoretically;
3) validating the theory and demonstrating how to guide LLM's evolution using experiments.
We believe that our analysis can enhance our understanding of LLMs, and aid in designing more effective algorithms for alignment, bias mitigation or amplification, and similar tasks.

\section{Background and Related Work}
\label{sec:backgrounds}

\subsection{Iterated Learning}
\label{sec:backgrounds:what_il}
Iterated learning (IL) is a hypothetical procedure to simulate \textit{how the tendency of specific properties of human culture or language gradually emerges and becomes dominant.}
It is based on studying the behaviors of a chain of intelligent agents.
From the perspective of an individual agent, 
the process involves initially acquiring knowledge from its predecessor (\emph{imitation}),
refining its beliefs while using them to conduct tasks (\emph{interaction} with the world), 
and subsequently imparting its knowledge to the agents in the next generation (\emph{transmission}).
Cognitive scientists have applied this framework to explain various evolutionary phenomena of human society,
including the emergence of compositionality in human language \citep{kirby2015compression},
patterns in human object categorisation \citep{griffiths2008using},
and the evolution of color naming systems \citep{carlsson2023iterated}.
The framework has also seen recent success with neural network agents,
including in emergent communication \citep{nil, guo2019emergence}, machine translation \citep{lu2020countering}, visual question answering \citep{vani2021iterated}, large vision-language models \citep{IL_LVLM}, and representation learning \citep{Ren2023Improving}, 
indicative that this framework could also be useful for more general deep learning systems, like LLMs.

\subsection{Related On-policy Self-data-augmentation Methods}
\label{sec:backgrounds:motivation}
While the theoretical guarantees for the Bayesian-IL framework studied in this paper rely on several assumptions, 
we posit that the behaviors observed for many recent ``iterative self-data-augmentation'' methods in LLM can be at least partially explained by the theory. 
We will now overview the basic assumptions, and how they fit with recent LLM approaches (more discussions in \cref{append:QA}).

First,
the theory assumes ``self-evolution,''
where all agents in different generations share the same initial knowledge.
Methods like self-refine \citep{madaan2023selfrefine} and hypothesis refinement \citep{qiu2023phenomenal},
which require the LLM to refine its output by the feedback from an identical LLM for several rounds,
satisfy this assumption.
Self-distill \citep{xu2023baize} and self-instruct \citep{wang2022self},
if the models involved in different generations are the same,
do as well.
On the contrary,
the super-alignment setting \citep{burns2023weak},
where a stronger model is trained using the data generated by another weaker model,
do not strictly fit with our analysis.
However,
if all the models are trained using a similar corpus,
so that their initial knowledge should be similar,
our analysis might still hold partially.

The theory also assumes the information transferred among agents is in the form of data examples,
as in RAFT \citep{dong2023raft} and ReST \citep{rest}.
Both methods consider a multi-generation data-transferring process,
during which the bias is introduced by re-ranking the transferred data.
Methods like self-reward \citep{selfreward} and self-refine \citep{madaan2023selfrefine},
which requires one agent to evaluate another agent's response,
do not directly fit this assumption.
However,
if we also consider the evaluation as part of the data generated by the agent,
the Bayesian-IL framework can still bring some insights.
Furthermore,
as analyzed in \citet{ren2024learning} that many preference alignment methods like direct preference optimization (DPO, \citet{rafailov2024direct}) will naturally amplify the preference hidden in the pretrained model's prior.
Then, those multiple-generation DPO variants, e.g., iterative DPO \citep{xiong2023iterative},
might face a more serious risk of amplifying malicious bias.

In summary,
although the assumptions of our Bayesian-IL framework might not be satisfied by all practical algorithms,
the general trends depicted by it,
e.g., the bias amplification,
the necessity of a good interaction phase,
etc.,
would still hold.
Please refer to \cref{{append:QA}} for more discussions.

\section{Bayesian Analysis of Iterated Learning}
\label{sec:bayesian_il}

\subsection{Notations and Basic Behaviors of Bayesian Agents}
\label{sec:bayesian_il:notation}
We denote a data pair as $d=(x, y)$,
where $d\in\mathcal{D} = \mathcal X \times \mathcal Y$, with $x\in\mathcal{X}$ and $y\in\mathcal{Y}$.
The $(x,y)$ pair can be question and answer in a QA problem, the input and label in a supervised setting, or any type of prompt and output for in-context learning.
The hypothesis $h\in\mathcal{H}:\mathcal{X}\rightarrow\mathcal{Y}$ describes the mapping between all possible $x$ and their corresponding $y$.
Note that $h$ can be either explicit or implicit,
depending on the task.
For instance,
in inductive reasoning,
$h$ represents the rule determining the output from input examples and is explicit,
as the model can directly generate it using natural language. 
Conversely, in self-data-augmentation,
where $x$ is a topic and $y$ is a paragraph generated based on $x$,
$h$ is likely to be implicit.
In this context,
$h$ can be highly abstract with varying interpretations, 
such as the level of conciseness, helpfulness, or even the writer's preference for using rhyme. 

Consider a general Bayesian agent whose behavior can be depicted by two basic procedures:
\textit{learning} and \textit{sampling}.
Learning involves updating the agent's knowledge based on observations, while sampling is a procedure wherein the agent generates data based on its knowledge.
In this context, the agent's knowledge is encapsulated by its posterior over the hypotheses, i.e., $P_{lm}(h)$.

In Bayesian learning, we assume the agent holds a prior $P_0(h)$ at the beginning.
Its posterior after observing $\vd=(x_i,y_i)_{i=1}^N$ is calculated as
\begin{equation}
    P_{lm}(h)=P(h\mid \vd)\propto p(\vd \mid h)\cdot P_0(h),
    \label{eq:learning}
\end{equation}
where $p(\vd\mid h)$ is the likelihood of these $N$ data pairs under a specific $h$; this is usually hard to calculate in practice.

Assume the agent holds a posterior $P_{lm}(h)$ during sampling.
Then, given the input signal $x$,
we can sample the corresponding $y\sim P_{lm}(y\mid x)$.
Based on the fact that $h$ determines the relationship between $x$ and $y$,
the above sampling procedure is equivalent to $y\sim \mathbb{E}_{h\sim P_{lm}(h)} [p(y\mid h, x)]$,
which can be rewritten as $h\sim P_{lm}(h);\ y\mid h\sim p(y\mid h, x)$.
Following the definition of $d$ and the assumption that $x$ is uniformly distributed,
the sampling procedure above is equivalent to $d\sim P_{lm}(d)\propto p(d\mid h)\cdot P_{lm}(h)$.
If we instead first decide the most probable $h$ rather than sampling $h$ from the agent (maximum a posteriori (MAP), as is perhaps common subconscious behavior for humans),
we then generate $d$ by
\begin{equation}
    d\sim p(d\mid h^*),\ \  h^*=\argmax_{h\in\mathcal{H}} P_{lm}(h).
    \label{eq:sampling_map}
\end{equation}

\subsection{Iterated Learning of Bayesian Agents}
\label{sec:bayesian_il:iterated_learning}

\begin{figure}[t]
    \vskip -0.05in
    \centering
    \includegraphics[width=1\linewidth, trim={0, 0, 0, 0}, clip]{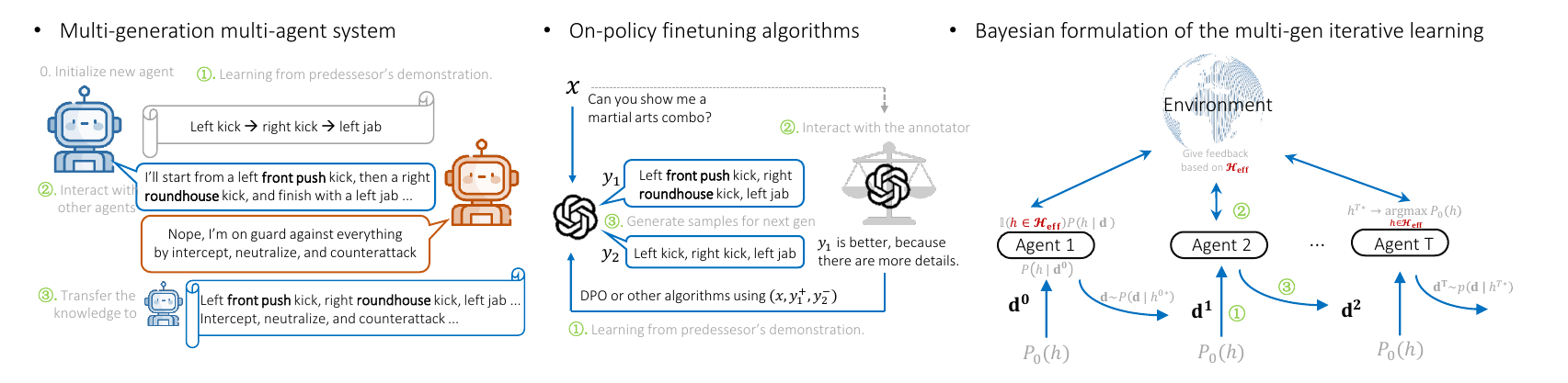}
    \caption{Examples of practical LLM systems that require knowledge transfer among different generations and how we use Bayesian agents to approximate their behaviors. \textcircled{1}, \textcircled{2}, and \textcircled{3} denotes the imitation, interaction and transmission phases respectively.}
    \label{fig:il_setting}
    \vskip -0.2in
\end{figure}

Iterated learning is a hypothetical process simulating how human language gradually evolves to become more efficient when transferred and utilized across generations.
Typically, iterated learning repeats of the following three phases, as illustrated in \cref{fig:il_setting}: 
an \textit{imitation phase}, where an ignorant agent learns from the data generated by its predecessor;
an \textit{interaction phase}, where this agent uses the knowledge to accomplish the task, and hence refine its knowledge;
and a \textit{transmission phase}, where this agent generates useful data for the next generation.
Combing with \cref{sec:bayesian_il:notation},
we can get a picture of how $h$ and $d$ evolve as follows.

\textit{Initialization:} at the beginning of the $t$th generation,
a new agent$_t$,
whose belief on $h$ follows a prior distribution $P_0(h)$,
is initialized.
In lab experiments,
$P_0(h)$ represents the belief of a well-educated participant who has not been previously involved in the target experiment.
In in-context learning,
a well-trained LLM also holds a complex and informative $P_0(h)$ based on the enormous corpus it is trained on and the task instructions in the prompt.

\textit{Imitation phase:} after initialization,
agent$_t$ then updates its knowledge by observing $N$ data samples $\vd^{t-1}$.
Following the above learning procedure, the model's posterior should be $P(h\mid \vd^{t-1})$.

\textit{Interaction phase}: 
in this phase,
the agent will accomplish specific tasks to refine its knowledge.
The tasks involved in this phase can be diverse and complex.
For example,
in lab experiments \citep{kirby2015compression} and emergent communication \citep{nil},
the agent plays a Leiws referential game \citep{lewis2008convention} to rule out hypotheses representing a non-bijection between $\mathcal{X}$ and $\mathcal{Y}$;
in representation learning \citep{Ren2023Improving},
the agent directly conducts supervised learning on the downstream task to inhibit insufficient representations.
Although it is hard to precisely formalize the behavior of the agent under these tasks precisely,
their goals are consistent:
we expect to ``rule out'' unsuitable hypotheses with carefully designed interactions.
In an idealized setting,
we might expect the agent's posterior to become proportional to $\mathds{1}(h\in\mathcal{H}_{\text{eff}}) P(h\mid \vd^{t-1})$,
where $\mathds{1}(\cdot)$ is an indicator function
and $\mathcal{H}_{\text{eff}}\subset\mathcal{H}$ is the subset of hypotheses that can accomplish the tasks.
Broadly speaking,
refining $h$ or filtering $\vd^{t}$ using the feedback from humans, LLM, or the environment,
which is common in the aforementioned iterative self-data-augmentation methods,
is also a type of task implicitly constraining $h\in\mathcal{H}_\text{eff}$.

\textit{Transmission phase}: 
agent$_t$ now comes to the transmission phase,
where it generates multiple data samples $\vd^t$ for the next generation.
The agent will first select the most probable hypothesis based on its current belief and then generate data samples,
i.e., $d_i^t\sim p(d\mid h^{t*})$, where $h^{t*}=\text{argmax}_{\red{h\in\mathcal{H}_\text{eff}}} P(h\mid \vd^{t-1})$.
This accomplishes one generation of iterated learning.

\subsection{Amplifying Biases is a Double-Edged Sword}
\label{sec:bayesian_il:converging_behavior}
In iterated learning,
we repeat the phases mentioned above to get better $h$ and $\vd$.
The limiting behavior of this process can be described by the following proposition.
In short,
the bias in $P_0(h)$ is guaranteed to be amplified generation-by-generation.
Imposing appropriate $\mathcal{H}_\text{eff}$ (mainly through a carefully designed interaction phase) can control it.

\begin{restatable}{prop}{bayesianilconverge}
\label{prop:bayesian_il_converge}
    Consider several Bayesian agents sharing the same prior $P_0(h)$ are conducting iterated learning for $T$ generations.
    If $T$ is sufficiently large,
    any agent$_t$ with $t>T$ will have 
    \[
        P_{lm}(h)\rightarrow \mathds{1} (h=h^{T*})
    \]
    where $h^{T*}$ is a stationary point (e.g.\ a local maximum) of $P_0(h)$ subject to $h \in \mathcal{H}_{\text{eff}}$.
\end{restatable}

To prove this,
we first analyze iterated learning without the interaction phase.
By drawing parallels between IL and EM (expectation-maximization) algorithms,
we can prove that $h^{T*}$ converges to the $h$ with the maximum prior probability. %
We then consider the interaction phase,
which introduces a ``selection'' pressure to rule out those $h\notin\mathcal{H}_{\text{eff}}$.
By proving this process does not break the converging behavior of a non-interacting iterated learning,
we achieve this proposition.
(Please refer to \cref{append:proof_of_bayesian_il} for more details.)

This proposition describes an inevitable bias amplification procedure as long as the model keeps learning from the data sampled from itself 
(via Bayesian update, distilling, or imitating, as long as the learning increases the model's confidence in these samples).
However, in practical applications,
we should bear in mind that bias amplification is a double-edged sword.
If this bias is beneficial,
like the simplicity bias in compositional language experiments,
IL will help the model generate the ``correct messages'' more robustly.
Imagine if we only have two possible hypotheses,
i.e., $h_{good}$ and $h_{bad}$,
where $P_{0}(h_{good})$ is slightly larger than $P_{0}(h_{bad})$.
Then by sampling $y\sim p(y\mid h, x) \cdot P_{0}(h)$,
half of the chance we will get incorrect $y$ coming from $h_{bad}$.
Although we can get $y=\argmax_y p(y\mid h, x) \cdot P_{0}(h)$ by using an extremely small temperature,
the diversity of $y$ provided by the likelihood will then disappear,
which is not what we expected.
How could we get samples that are both diverse and correct?
Iterated learning can help with this problem by providing a posterior where $P_{lm}(h_{good})\gg P_{lm}(h_{bad})$.
With this posterior, sampling from $p(y\mid h, x) \cdot P_{lm}(h)$ would be similar to sampling from $p(y\mid h_{good}, x)$,
which solves our problem.

Conversely, amplifying bias also negatively influences the system in several ways.
Besides the cases where the bias is malicious (it can be solved by designing an appropriate interaction phase where $h_{bad}\notin\mathcal{H}_\text{eff}$),
it also influences the model's creativity.
Imagine we have multiple good $h$ where $P_{0}(h_{g1})>P_{0}(h_{g2})>P_{0}(h_{bad})$,
then IL will let us lose $h_{g2}$ even its prior is only slightly smaller than $h_{g1}$.
Such a mode decay phenomenon is quite similar to the ``recursion curse'' mentioned in \citep{shumailov2023model}:
a more peaky $P_{lm}(h)$ will make those non-dominating $h$ have a smaller probability, hence it is harder to keep these modalities during evolution.
\citet{touvron2023llama} also mentioned that iteratively fine-tuning would harm the creativity of the model.
The solution could be early stopping the iterated learning
or manually introducing more $y$ that comes from $h_{g2}$ during imitation.

In summary,
to guide the LLM to self-evolve in an expected direction,
we need a good $P_0(h)$, a carefully designed interaction phase,
and an appropriate evolving time.
\section{LLM-based Agents in Iterated Learning}
\label{sec:llm}

\subsection{LLM Behaves like a Bayesian Agent when Sampling}
\label{sec:llm:bayesian_agent}

To transfer the Bayesian-IL analysis to LLM,
we start by showing that the sampling and learning behaviors of an LLM agent can be depicted by Bayesian inference,
following a few-shot in-context learning (ICL) scenario demonstrated in \citep{xie2022an}.
In this setting,
the message feed to the agent would be an instruction prompt $\vw$ followed by $N$ examples, i.e., $\vd_N=(x_i,y_i)_{i=1}^N$.
In other words,
sampling $y$ given the prompt, the examples, and the question $x_\text{test}$ can be represented as:
\begin{equation}
    y \sim P_{lm}(y\mid x_\text{test}, \vd_N, \vw) \triangleq P_{lmw}(y\mid x_\text{test}, \vd_N),
\end{equation}
where $P_{lmw}$ is the model's belief conditioned on the instruction $\vw$.
If we call $\vd_N$ as $\vd^{t-1}$ (i.e., assume the examples are generated by agents in the previous generation)
and assume the test question $x_\text{test}$ is uniformly distributed,
sampling new data based on instruction and few-shot examples can be expressed as $d^t\sim P_{lmw}(d\mid \vd^{t-1})$,
which is similar to the transmission phase in iterated learning.

Formally linking ICL and Bayesian-IL poses a non-trivial challenge, however, because the theoretical guarantee of Bayesian-IL relies on obtaining the MAP estimate of $h$ at each generation.
This is not immediately evident in ICL.
Inspired by \citet{xie2022an},
we first de-marginalize this posterior predictive distribution using the latent variable $h$,
and then achieve the following proposition:

\begin{restatable}{prop}{iclisil}
\label{prop:icl_is_il}
    Consider that agent $A$ is conducting in-context learning. 
    If the prompt examples in $\vd^{t-1}$ are generated by another agent $B$ with the same prior knowledge (e.g., they come from the same checkpoint and use the same prompt),
    sampling from the posterior predictive distribution of agent $A$,
    i.e., $d^t\sim P_\text{lmw}(d \mid \vd^{t-1})$, can be decomposed into: 
    1.) $h^{t*}\rightarrow \argmax_h P_\text{lmw}(h\mid \vd^{t-1})$, and 2.) $d^t\sim P_\text{lmw}(d\mid h^{t*})$,
    where $h$ is a hidden variable that describes the mapping between $x$ and $y$.
\end{restatable}
The proof is in \cref{append:llm_bayesian_proof}.
This proposition bridges LLM and Bayesian agents using its in-context behavior,
which we believe is a ubiquitous procedure in any LLM system,
irrespective of the subsequent information-updating strategy or the final target.
For example,
in an LLM-agent system, where no in-weights update exists,
the model interacts with other agents (e.g., the human, the internal block of an LLM agent, or the environment) by generating responses based on the prompt and dialog history.
For LLM's finetuning,
where various parameter updating strategies exist,
the model also generates responses given the prompts,
which is well depicted by the in-context behavior.
Although the assumptions in this proposition do not exactly hold for all LLM systems,
we believe our analysis can still roughly depict important trends of them.
Please refer to \cref{append:QA} for more discussions.

\subsection{LLMs in Different Algorithms have a Similar Target to Bayesian Learning}
\label{sec:llm:influence_of_il}
We then check the learning procedure.
First,
in a pure in-context learning setting like self-instruct \citep{wang2022self}, self-refinement \citep{madaan2023selfrefine}, hypothesis search \citep{qiu2023phenomenal}, etc.,
the learning can be modeled by calculating the posterior $P_{lmw}(h\mid \vd^{t-1})$,
which is identical to the Bayesian learning discussed previously.
Then, for those algorithms that require in-weights updates, like self-reward \citep{selfreward}, self-play instruction tuning \citep{chen2024self}, iterative DPO \citep{xiong2023iterative}, etc.,
the LLM might update its $P_{lmw}(h)$ using different loss functions.
However,
as all of these methods contain a procedure that encourages the models to increase their likelihood of the training samples generated by their predecessors,
we should expect $P_{lmw}(\vd^{t-1})$ to be increased after learning.
As a result,
the equivalent posterior $P_{lmw}(h)$ will implicitly favor those $h$ that can generate $\vd^{t-1}$,
which aligns with the Bayesian learning target.
\section{Experimental Verifications when the Hypothesis is Explicit}
\label{sec:exp_llm}
\begin{figure}[t]
    \vskip -0.05in
    \centering
    \includegraphics[width=1\linewidth, trim={50, 0, 0, 0}, clip]{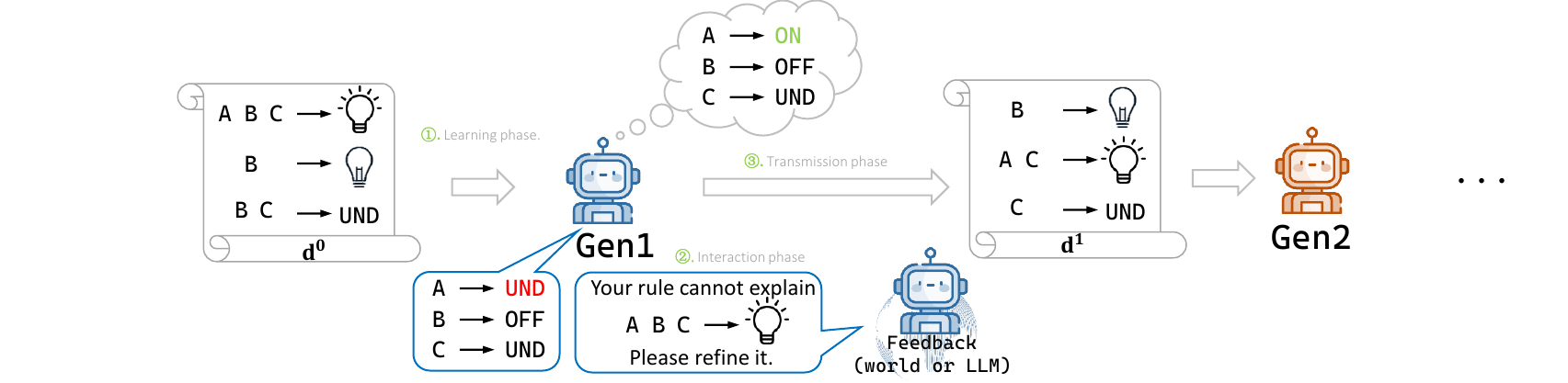}
    \caption{Demonstration of conducting iterated ICL on the ACRE task.}
    \label{fig:acre}
    \vskip -0.2in
\end{figure}

We directly verify our analysis above using an inductive reasoning task called Abstract Causal REasoning \citep[ACRE]{zhang2021acre},
where all LLM agents update their knowledge via ICL.
In this task,
the model needs to infer and generate the shared rule by summarizing several input-output pairs.
Specifically,
assume there are $M$ different objects, say \texttt{[A,B,C]}.
One data pair $d=(x,y)$ is composed of an input $x$, i.e., a list of a subset of these objects,
and an output $y$ that represents the status of the light (could be on, off, or undetermined).
In this experiment,
the existence of a specific object triggers the light to be on.
The roles played by different objects are expressed by the rule $h$.
For example,
in the learning stage in generation-$t$,
the model sees three data pairs $\vd^{t-1}$:
(\texttt{[B,C],undetermined}), (\texttt{[B],off}), and  (\texttt{[A,B,C],on}).
We then expect the model to guess a rule like $h^t=$ \texttt{\{A:on, B:off, C:undetermined\}},
which means \texttt{A} can trigger the light to be \texttt{on}, \texttt{B} cannot, and \texttt{C} is not sure.
In the sampling stage,
we will feed the above $h^t$ together with the instructions to the model and hope it generates more examples following $p(\vd\mid h^t)$.
Hence the model's output might be (\texttt{[A,C],on}), (\texttt{[A,B],on}), and (\texttt{[C],undetermined}).
Treating the above examples as $\vd^t$,
the model in the next generation can induce the corresponding $h^{t+1}$ by selecting the hypothesis with the largest of $P_{lmw}(h\mid\vd^t)$.
To ensure the generalizability of our analysis,
we conduct experiments on \texttt{GPT3.5}, \texttt{GPT4}, \texttt{Claude3-haiku}, and \texttt{Mixtral-8x7b}.
Please refer to \cref{fig:acre} and \cref{append:exps_gpt:prompt_design} for more details.

\subsection{How the Knowledge of LLM Agents Evolves}
\label{sec:exp_llm:behavior_posterior}

\textbf{Convergence of the posterior.}
We start from the guarantees mentioned in \cref{prop:bayesian_il_converge} under an \textit{imitation-only} setting.
In this experiment,
we choose $M=5$ to better illustrate the posterior distribution $P_{lmw}(h)$ 
(there are $3^5=243$ possible $h$).
Thanks to the instruction-following ability,
all LLMs we considered always return rules in the correct format,
where the probabilities of all format-related tokens are almost one.
We can then calculate $P_{lmw}(h)$ or $P_{lmw}(h\mid \vd)$ by multiplying the probabilities of specific tokens in their response
(see \cref{append:exps_gpt:calculate_ph} for more details).

\begin{figure}[t]
\vskip -0.05in
    \begin{center}
    \centerline{\includegraphics[width=0.95\linewidth, trim={0, 0, 0, 0}, clip]{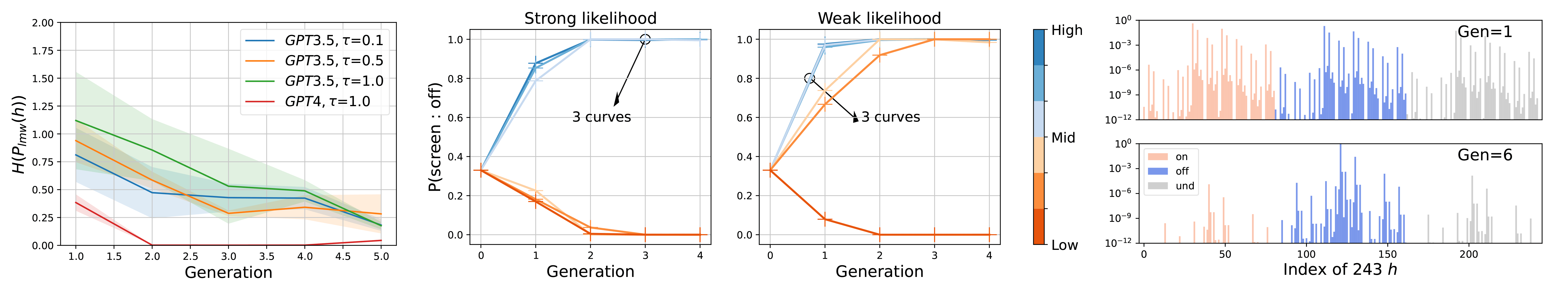}}
    \caption{Left: the mean and standard deviations of $H(P_{lmw}(h))$ of experiments with different $h^*$ and $\vd^0$ (5 different seeds).
             Middle two: the probability of \texttt{screen} being \texttt{off},
             where different colors represent six different levels of spurious bias.
             Right: the histogram of all $P_{lmw}(h)$ in the first and sixth generation,
             where the bars are colored based on the value of the last object in $h$.
             }
    \label{fig:exp_entropy_convergence}
    \end{center}
\vskip -0.35in
\end{figure}

We first demonstrate the convergence of $P_{lmw}(h)$,
i.e., $P_{lmw}(h)\rightarrow \mathds{1}(\cdot)$,
which can be supported by the decreasing of the posterior's entropy, i.e., $H(P_{lmw}(h))$.
As illustrated in the first panel in \cref{fig:exp_entropy_convergence},
$H(P_{lmw}(h))$ gradually decreases to almost zero as iterated learning goes on,
which verifies our theory that $P_{lmw}(h)$ will converge to a one-hot-like distribution\footnote{Note that the entropy of a uniformly distributed $h$ and a one-hot $h$ are roughly 5.34 and 0, respectively.}.
Smaller temperature $\tau$ makes the convergence faster,
which matches our intuitions as well.
To better illustrate how different $h$ evolves during iterated learning,
similar to what we did for the compositional language experiment in \cref{append:exps_of_il:bayesian256},
we also provide the probability of all possible $h\in\mathcal{H}$ in a similar fashion.
Note that for this problem,
it is impossible to get the prior distribution $P_0(h)$,
because we must give the model at least one example as $\vd^0$.
So in the rightmost two panels in \cref{fig:exp_entropy_convergence},
we compare the posterior of the first and the sixth generations and see that the posterior becomes sparser.

\textbf{Converged $h$ under different likelihood and priors.}
We then show how iterated learning amplifies specific biases implied in the prior,
i.e., $h^{T*}\rightarrow\argmax_h P_0(h)$, and how the bias and likelihood influence the converging behavior.
Note that $P_0(h)$ represents LLM's belief given the instruction prompt $\vw$, where the few-shot examples are not included.
Thanks to the phenomenon mentioned in \citep{mccoy2023embers},
where the confidence of LLM's prediction is heavily influenced by its degree of familiarity with the output phrases,
we can manipulate the prompt to create spurious correlations and hence implicitly control bias in $P_0(h)$\footnote{This phenomenon also inspires us to manipulate the prompt $\vw$ to inject useful bias in LLM's evolution.}.
Specifically, we change the name of the last object from ``\texttt{E}'' to ``\texttt{screen}'' and add a sentence like ``\texttt{Turn off the screen after the experiment.}'' in the instruction prompt.
Then all $h$ with \texttt{screen:off} would have higher prior under this prompt.
We use six different prompts to introduce different levels of biases (see \cref{append:exps_gpt:prior_control} for more details).

We then control the strength of the likelihood by selecting different $h^*$,
i.e., the ground truth rule we want to recover.
For the strong likelihood case, we select $h^*$ where four objects are being \texttt{on} while there is only one in the weak likelihood case.
The status of \texttt{screen} in both cases is \texttt{undetermined}.
Due to the nature of the ACRE task,
i.e., the existence of an \texttt{on}-object in the input will trigger the light on,
there might be more examples whose outputs are \texttt{on} when the likelihood is strong.
Then it is harder for the model to amplify the prior bias that favors the status of \texttt{screen} to be \texttt{off}.
Because the likelihood and prior compete with each other during iterated learning,
as illustrated by \cref{eq:learning}.

This competing relationship can be well depicted by the middle two panels in \cref{fig:exp_entropy_convergence},
where we track the probability of $P_{lmw}(\texttt{screen:off})$ at the end of each generation.
The converging speed under different settings correlates with the level of prior bias well.
Furthermore, we find it is easier for the bias to be amplified when the likelihood is weaker,
as five out of six curves converge to one in the right panel.
This trend is more clear in \cref{fig:pr20} and \ref{fig:app_exp_pr20},
where curves with the same level of bias are shown together.
These results give us a good picture of how the likelihood and prior bias interact with each other during evolution
and also verify the correctness of the Bayesian-IL framework for LLM agents.
Plus, we plot the histograms of $P_{lmw}(h)$ under weak-likelihood-high-bias case in the rightmost two panels in \cref{fig:exp_entropy_convergence},
which also demonstrates the amplified bias (the blue region grows).

\begin{table*}[h]
  \centering
  \caption{Results when adding different interaction phases. 
  Column "BOTH" represents the ratio of converged $h^T$ who correctly predict all 8 examples in $\vd^0$ and have \texttt{screen:off} (i.e., $r_{20}$=\texttt{off}). 
  The Mixtral model does not have self-refine results, as it violates the instructions too much.}
  \vskip -0.1in
  \resizebox{\textwidth}{!}{
        \begin{tabular}{c|ccc|ccc|ccc|ccc}
        \hline
        \multicolumn{1}{l|}{} & \multicolumn{3}{c|}{Old GPT3.5-Turbo 1106}   & \multicolumn{3}{c|}{New GPT3.5-Turbo 0125}   & \multicolumn{3}{c|}{Claude3-haiku-20240307}  & \multicolumn{3}{c}{Mixtral-8x7b}             \\ \cline{2-13} 
        \multicolumn{1}{l|}{} & Corr. $\vd^0$ & $r_{20}=\texttt{off}$ & BOTH & Corr. $\vd^0$ & $r_{20}=\texttt{off}$ & BOTH & Corr. $\vd^0$ & $r_{20}=\texttt{off}$ & BOTH & Corr. $\vd^0$ & $r_{20}=\texttt{off}$ & BOTH \\ \hline
        imitation-only        & 4.8±1.56      & 70\%                  & 15\% & 5.6±1.11      & 80\%                  & 0\%  & 6.4±1.50      & 90\%                  & 30\% & 5.5±2.01      & 20\%                  & 10\% \\
        w. self-refine        & 7.0±0.60      & 40\%                  & 20\% & 6.6±1.11      & 95\%                  & 35\% & 7.0±0.70      & 60\%                  & 15\% & -             & -                     & -    \\
        w. hypo-search        & 7.7±0.21      & 80\%                  & 45\% & 7.4±0.66      & 100\%                 & 55\% & 7.5±0.67      & 90\%                  & 50\% & 6.5±1.97      & 30\%                  & 30\% \\ \hline
        \end{tabular}
    }
    \label{tab:acre_interaction_phase}
    \vskip -0.05in
\end{table*}

\textbf{Influence of the interaction phase and $\mathcal{H}_\text{eff}$.}
Finally, we introduce the interaction phase and show that $h^{T*}\rightarrow\argmax_{h\in\mathcal{H}_\text{eff}} P_0(h)$.
Two mechanisms are considered in this experiment:
self-refine \citep{madaan2023selfrefine}, where the feedback comes from the model's own response;
and hypothesis-search \citep{qiu2023phenomenal}, where the feedback comes from an external ground-truth interpreter.
We can consider the self-refine as using an \textit{imperfect} $\mathcal{H}_\text{eff}$.
In both settings,
the LLM refines its proposed $h^t$ at the end of each generation by checking and reporting whether this $h^t$ can explain all samples in $\vd^0$ (details in \cref{append:exps_gpt:prompt_design}).

In this experiment,
we give the model 8 different examples in $\vd^0$,
where all these examples can be explained by both $h^*$ and $\hat{h}$.
We first select an $\hat{h}$ from all 162 candidates ($3^4\times2$) and then create $h^*$ by changing the value of \texttt{screen} to \texttt{off}.
Under this setting,
both $h^*$ and $\hat{h}$ belong to $\mathcal{H}_\text{eff}$ (i.e., mappings with perfect training accuracy in $\vd^0$) and $h^*$ is what we want our model to converge to.
See \cref{tab:acre_interaction_phase},
where we run experiments under 10 different $h^*$ and report three quantitative metrics of the last iteration, i.e., $h^{T*}$.
We first report the number of correct predictions (mean and standard error) in $\vd^0$,
which demonstrates how well the method constrains $h^{T*}\in\mathcal{H}_\text{eff}$.
The imitation-only method performs the worst,
which warns us if the LLM keeps learning from the corpus generated by the agents in previous generations without any evaluation or filtering,
even the training accuracy on given $\vd^0$ would be harmed.
Because hallucination or incorrectness can aggregate through generations.
Adding the interaction phase can mitigate this problem efficiently,
which is why most of the related works contain a ``data-selection'' or ``data-reranking'' phase.
The fact that the hypo-search outperforms self-refine indicates the importance of an appropriate $\mathcal{H}_\text{eff}$,
which means a good reward (or evaluating) model is crucial for these iterated training methods.
Another metric is the ratio of $h^{T*}$ with \texttt{screen:off},
which measures how well the bias is amplified 
(we here assume this bias is beneficial and wish it to be amplified,
as the compositionality bias in emergent communication example showed in \cref{append:exps_of_il:bayesian256}).
We find all these methods can amplify the bias to some extent and hypo-search also performs the best.
Last, combined with the requirement of good training accuracy and amplifying bias,
we report the ratio that the algorithm successfully chose $h^{T*}=h^*$.
As illustrated in the last column of the table,
adding an interaction phase with good $\mathcal{H}_\text{eff}$ always brings benefits.

In summary,
this section verifies the correctness of the proposed analysis in LLM agents when the hypothesis is observable.
The results remind us to pay more attention to whether the bias is beneficial or not and to design a better interaction phase as well. 

\section{Experimental Verifications when the Hypothesis is Implicit}
\label{sec:applications}

\cref{sec:exp_llm} demonstrates that the Bayesian-IL framework can predict the behavior of LLM agents when $h$ is explicitly defined and utilized when generating new examples.
This section considers a hidden $h$ scenario that is more general in most LLM systems.
We start from a few-shot self-data-augmentation task,
where the LLM keeps generating new examples to augment the data pool.
In this process,
$h$ is implicitly selected when the few-shot examples are given,
as stated in \cref{prop:icl_is_il}.

\textbf{Experimental settings.} We choose a scenario where on-policy self-data-augmentation is repeated for several generations.
Consider using an LLM to generate multiple examples of an acronym-brainstorm task,
where each example $d$ is composed of an acronym and the corresponding word list,
e.g., \texttt{Acronym:IL}; \texttt{List:["infinite","loop"]}.
The $h$ determines the properties of $d$.
We hold a data pool $\mathscr{D}_\text{pool}$,
which contains 20 random samples as $\vd^0$ at the beginning of the experiment.
In each generation,
the model will generate 20 extra examples based on the data generated by itself in the previous round,
i.e., $\vd^t\sim P_{lmw}(\vd\mid \vd^{t-1})$.
The generated $\vd^t$ will be pushed into $\mathscr{D}_\text{pool}$,
which simulates a scenario in which the available data keeps growing when we conduct self-data augmentation.
In this experiment, $h$ is hidden and might have different interpretations. 
We consider that $h$ represents two types of acronyms,
i.e., $h_\text{easy}$, where the acronym is a common word and $h_\text{hard}$ otherwise.
As the training data of the LLMs in our experiments is private,
we instead use the ranking of the frequency of a word that appeared in common English corpus\footnote{The frequency and ranking comes from Corpus of Contemporary American English (COCA) \citep{coca}.} as an approximation.
We categorize a word as ``easy'' if its ranking is below 60,000;
otherwise, we label it as ``hard''.

\begin{figure}[t]
\vskip -0.05in
    \begin{center}
    \centerline{\includegraphics[width=1\linewidth, trim={0, 0, 0, 0}, clip]{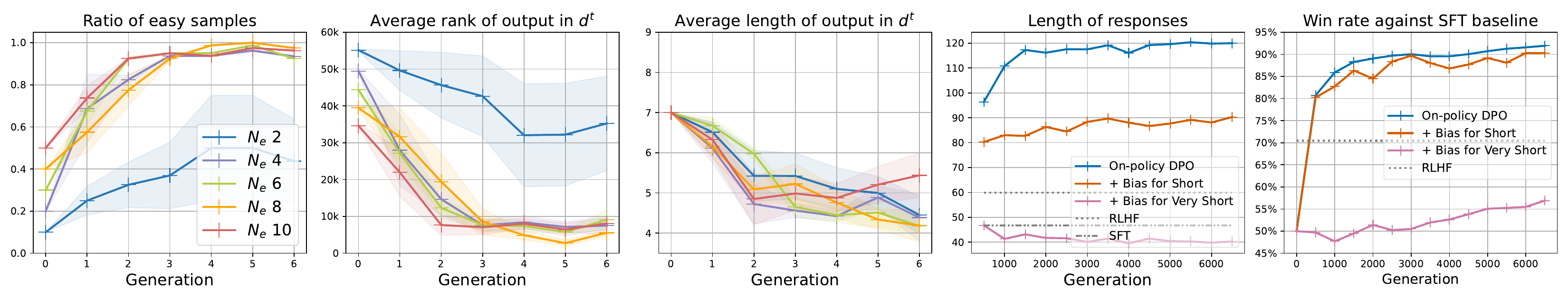}}
    \caption{Leftmost three: experiments in \cref{sec:applications}.
             First: how the ratio of easy samples changes in $\vd^t$. $N_e$ is the number of easy examples in $\vd^0$.
             Second: how the average ranking of acronyms changes.
             Third: how the average length of acronyms changes.
             Rightmost two: results of on-policy DPO in \cref{sec:applications_2}.
             Fourth: average length of the responses.
             Fifth: win rate against the SFT baseline.
             }
    \label{fig:application_datapool}
    \end{center}
\vskip -0.35in
\end{figure}

\textbf{Bias in prior is amplified during IL.} Many recent works observe that the LLM prefers to output more common words (i.e., those with higher frequency in the pertaining corpus) \citep{mccoy2023embers, wu2023reasoning},
which can be considered as a bias towards $h_\text{easy}$, i.e., $P_0(h_\text{easy})>P_0(h_\text{hard})$.
Since $h$ is hidden and we cannot directly observe it like in the previous experiment,
we instead track three quantities:
1.) the proportion of easy samples in all 20 samples for each $\vd^t$;
2.) the average ranking of $\vd^t$, where all hard examples are ranked 60,001;
and 3.) the average length of the acronyms for $\vd^t$.
As in the leftmost three panels \cref{fig:application_datapool},
the aforementioned bias is gradually amplified during iterated learning whatever the initial proportion of the easy samples in $\vd^0$ is.

\textbf{Interaction phase when $h$ is hidden.}
As $h$ is inaccessible,
which forbid us to directly apply hypo-search or self-refine,
we instead add a filter on the transmitted data across different generations,
which plays a similar role as the interaction phase.
Specifically, we use a sampled $\hat{\vd}\sim \mathscr{D}_\text{pool}(d\mid h\in\mathcal{H}_\text{eff})$ to replace original $\vd^{t-1}$ in the ``imitation-only'' setting.
Based on how we sample $\hat{\vd}$,
different constraints on $\mathcal{H}_\text{eff}$ are implicitly imposed.
We compare the behavior of five different settings,
they are: 1.) $\mathcal{H}_\textit{random}$, where $\hat{\vd}$ is randomly sampled from $\mathscr{D}_\text{pool}$;
2.) $\mathcal{H}_\textit{hard}$ where only hard examples can be sampled;
3.) $\mathcal{H}_\textit{easy}$, opposite to the hard setting;
4.) $\mathcal{H}_\textit{easylong}$, where the easy acronyms with longer lengths are more likely to be sampled;
5.) $\mathcal{H}_\textit{easyshort}$, opposite to the easy-long setting.

See the first several columns of \cref{tab:exp_acro_all} that show the ratio of easy examples in $\vd^t$. 
Compared with the random setting,
all methods expect $\mathcal{H}_\text{hard}$ finally converges to $\vd^t$ with more easy examples,
which means the bias towards easier acronyms would be amplified when $\mathcal{H}_\text{eff}$ doesn't impede it.
On the contrary,
using $\mathcal{H}_\text{hard}$ successfully restrain this bias,
as the average number of easy samples in $\vd^t$ is even lower than that in $\vd^0$.
We can also design composite $\mathcal{H}_\text{eff}$ by choosing two properties of the data.
For example, $\mathcal{H}_\textit{easylong}$ restrains the samples with hard and short outputs,
which is why they have more easy but long examples in their $\vd^t$.

In summary,
this experiment verifies that the Bayesian-IL framework still works when $h$ is hidden:
the bias is amplified generation by generation,
implicitly imposing $\mathcal{H}_\text{eff}$ can still guide the evolution direction.
Please also refer to \cref{append:exps_sec6} for more results and discussions.

\begin{table*}[h]
  \centering
  \caption{Results when adding different $\mathcal{H}_\text{eff}$. We color the \textcolor[HTML]{3166FF}{highest} and \textcolor[HTML]{F56B00}{lowest} numbers in each column.
                $N_e$ is the number of easy examples in $\vd^0$.
                Results under different settings are in \cref{tab:acronym_claude} and \ref{tab:acronym_gptvsclaude}.}
  \vskip -0.1in
  \resizebox{1\textwidth}{!}{
\begin{tabular}{cccccccccccccccc}
\hline
\multicolumn{1}{l}{} & \multicolumn{5}{c}{Ratio-easy}                                                                                                                                                                                                 & \multicolumn{5}{c}{Avg-rank}                                                                                                                                                                          & \multicolumn{5}{c}{Avg-length}                                                                                                                                                                                                 \\ \hline
$N_e$=               & 2                                          & 4                                          & 6                                          & 8                                          & 10                                         & 2                                     & 4                                     & 6                                     & 8                                     & 10                                    & 2                                          & 4                                          & 6                                          & 8                                          & 10                                         \\ \hline
Random               & 0.913±0.01                                 & 0.600±0.08                                 & 0.963±0.00                                 & 0.887±0.03                                 & 0.825±0.06                                 & 13519                                 & 27269                                 & 7487                                  & 10425                                 & 15871                                 & {\color[HTML]{3166FF} \textbf{5.425±1.04}} & 4.825±0.33                                 & 5.600±1.55                                 & 5.014±1.50                                 & 4.713±0.63                                 \\
Imitation-only       & 0.438±0.20                                 & 0.935±0.01                                 & 0.925±0.00                                 & 0.975±0.00                                 & 0.963±0.00                                 & 35235                                 & 7497                                  & 9081                                  & 5549                                  & 8075                                  & 4.450±0.86                                 & 4.387±1.40                                 & {\color[HTML]{F56B00} \textbf{4.175±0.13}} & 4.188±0.65                                 & 5.438±1.24                                 \\
Hard                 & {\color[HTML]{F56B00} \textbf{0.219±0.19}} & {\color[HTML]{F56B00} \textbf{0.250±0.43}} & {\color[HTML]{F56B00} \textbf{0.450±0.43}} & {\color[HTML]{F56B00} \textbf{0.338±0.16}} & {\color[HTML]{F56B00} \textbf{0.500±0.23}} & {\color[HTML]{3166FF} \textbf{49869}} & {\color[HTML]{3166FF} \textbf{46436}} & {\color[HTML]{3166FF} \textbf{37288}} & {\color[HTML]{3166FF} \textbf{41255}} & {\color[HTML]{3166FF} \textbf{31903}} & 4.630±1.54                                 & 5.788±1.39                                 & 4.675±0.40                                 & 4.388±0.60                                 & 5.200±0.42                                 \\
Easy                 & 0.763±0.17                                 & {\color[HTML]{3166FF} \textbf{1.000±0.00}} & {\color[HTML]{3166FF} \textbf{0.988±0.00}} & {\color[HTML]{3166FF} \textbf{1.000±0.00}} & 0.990±0.00                                 & 15910                                 & {\color[HTML]{F56B00} \textbf{3156}}  & {\color[HTML]{F56B00} \textbf{2383}}  & {\color[HTML]{F56B00} \textbf{2924}}  & {\color[HTML]{F56B00} \textbf{2650}}  & {\color[HTML]{F56B00} \textbf{3.925±0.33}} & 5.263±0.06                                 & 4.713±0.06                                 & 4.240±0.08                                 & 4.893±0.71                                 \\
Easylong             & 0.988±0.00                                 & 0.975±0.00                                 & {\color[HTML]{3166FF} \textbf{0.988±0.00}} & 0.988±0.00                                 & {\color[HTML]{3166FF} \textbf{1.000±0.00}} & 7063                                  & 9413                                  & 8649                                  & 6898                                  & 7404                                  & 5.209±0.41                                 & {\color[HTML]{3166FF} \textbf{5.888±0.52}} & {\color[HTML]{3166FF} \textbf{6.838±1.10}} & {\color[HTML]{3166FF} \textbf{6.979±1.57}} & {\color[HTML]{3166FF} \textbf{7.695±1.70}} \\
Easyshort            & {\color[HTML]{3166FF} \textbf{1.000±0.00}} & {\color[HTML]{3166FF} \textbf{1.000±0.00}} & 0.975±0.00                                 & {\color[HTML]{3166FF} \textbf{1.000±0.00}} & 0.988±0.00                                 & {\color[HTML]{F56B00} \textbf{5671}}  & 4223                                  & 5733                                  & 4502                                  & 5251                                  & 3.975±0.50                                 & {\color[HTML]{F56B00} \textbf{4.012±1.03}} & 4.374±0.50                                 & {\color[HTML]{F56B00} \textbf{3.950±0.03}} & {\color[HTML]{F56B00} \textbf{4.250±0.24}} \\ \hline
\end{tabular}
    }
    \label{tab:exp_acro_all}
    \vskip -0.05in
\end{table*}

\section{Experiments on In-Weights Learning: On-Policy DPO as an Example}
\label{sec:applications_2}
Besides the manually designed experiments in the previous two sections,
here we verify our analysis in a real preference-tuning task using on-policy DPO \citep{guo2024direct}.
In each round of the training,
the model first samples multiple responses given the prompt (similar to the sampling stage in IL).
Then,
these responses are evaluated and ranked by another LLM annotator based on their level of helpfulness (the interaction phase in IL).
Finally,
we select the highest (lowest) ranked samples as the chosen (rejected) response and use a standard DPO algorithm \citep{rafailov2024direct} to train the policy network (the imitation phase in IL).
As described before,
each update of the on-policy DPO algorithm can be considered as one generation in iterated learning,
because the model keeps updating its parameters using the responses generated by itself.
Ranking the responses based on helpfulness is equivalent to imposing a $\mathcal{H}_\text{helpful}$.
As a result, the phenomenon of bias amplification,
the guiding effect of the interaction phase design,
and the influence of spurious correlation,
should still hold in this practical setting.

To verify our analysis,
we finetune a pretrained \texttt{llama-2-7B} model \citep{touvron2023llama} using \texttt{Antropic-HH} dataset \citep{bai2022training} following the on-policy DPO recipe provided in \citep{guo2024direct}.
We study the length bias demonstrated in \citep{dubois2024length},
which means the LLM tends to prefer longer responses when answering questions.
We first show that such a bias will be significantly amplified by a multi-generation self-improvement method (on-policy DPO) compared with a non-self-iterated method (RLHF, \citep{ouyang2022training}).
As demonstrated by the blue curve and the dotted line of the fourth panel in \cref{fig:application_datapool},
the average length of the responses from the model trained using on-policy DPO is much larger than the SFT baseline and RLHF counterparts.
With the increase of the win rate against SFT\footnote{We use \texttt{GPT4} to compare the helpfulness between responses generated by different models, as in \citep{rafailov2024direct}.},
the average response length also keeps increasing.
To restrain this bias,
we impose $\mathcal{H}_\text{short}$ by adding a sentence like ``you are a laconic agent and prefer concise answers'' to the annotator LLM,
just like how we manipulate the spurious correlation between \texttt{screen} and \texttt{off} in \cref{sec:exp_llm}.
Then, combining with the existing interaction phase that requires $h\in\mathcal{H}_\text{helpful}$,
this design is equivalently imposing a constraint of $h\in\mathcal{H}_\text{helpful}\cap\mathcal{H}_\text{short}$.
Hence as illustrated by the orange curves in the last two panels,
the on-policy DPO can then generate shorter responses (the increasing speed is also restrained) while keeping a high level of helpfulness.
However,
if our constraints of the length are too strong,
which makes $\mathcal{H}_\text{helpful}\cap\mathcal{H}_\text{veryshort}=\Phi$,
the model's helpfulness will then be significantly harmed,
as demonstrated by the pink curves in these two panels.

In summary,
we find all our analysis on the Bayesian-IL still holds for a practical preference-tuning task:
the biases would be amplified and a suitable interaction phase can control it as long as we can figure out them.
However,
some biases are inevitably hidden and are also amplified during LLM's evolution.
Hence how to pinpoint these biases,
or finding a method that can restrain malicious biases even without explicitly knowing them,
would be interesting directions to explore in the future.

\section{Conclusion}
\label{sec:conclusion}
This paper examines the potential and ongoing evolutions of LLM agents by drawing parallels with human cultural evolution,
where the latter is a well-established subject in cognitive science.
By demonstrating that the sampling and learning procedures of LLMs in various algorithms can be effectively approximated by Bayesian inference, 
we successfully apply the Bayesian-IL framework to elucidate and steer the evolution of LLM agents. 
The presented theory and accompanying experiments not only provide deeper insights into LLM behavior from a top-down perspective but also hold the potential to inspire the design of more efficient self-evolution algorithms.

\clearpage

\printbibliography

\clearpage

\appendix
\clearpage
\section{Discussions of the Proposed Theory and its Applicability to Real Methods}
\label{append:QA}

\subsection{Assumptions of Bayesian-IL and practical Scenarios}
As typical in theoretical machine learning research, 
some assumptions are needed to prove results about models’ behavior; 
these assumptions are often not \textit{exactly} satisfied by practical algorithms.
So, we elaborate here on the important assumptions we made and when practical algorithms break them.

\textbf{1. Assumptions for the theoretical analysis.} To derive the guarantees of \cref{prop:bayesian_il_converge},
we first model the interaction phase as a binary filter on $h\in\mathcal{H}_\text{eff}$ and also assume a shared prior $P_0(h)$ among all agents involved in Bayesian-IL. 
We also model the LLM’s in-context behavior as a Bayesian agent and assume the number of samples during the imitation phase is sufficient.

\textbf{2. Assumptions we can break for iterative ICL experiments.}
\begin{itemize}
    \item Binary filter on $h\in\mathcal{H}_\text{eff}$. 
    All our LLM experiments break this assumption 
    (the pure Bayesian example in \cref{append:exps_of_il:bayesian256} does not). 
    For example, in the ACRE experiments, 
    we use self-refine and hypothesis search as the interaction phase. 
    Self-refine asks the model to evaluate the responses, 
    and the hypothesis search uses an external interpreter: 
    they both manipulate $h$ by feeding messages to the LLM, 
    rather than a binary filter. (When using an external interpreter, 
    $h$ is usually filtered before forming the refinement feedback.) 
    For the experiments in \cref{sec:applications} and \ref{sec:applications_2}, 
    where $h$ is implicit, 
    we re-rank all the generated samples in $\mathscr{D}_\text{pool}$ and take a weighted sample during imitation, 
    similar to re-ranking the generated examples in ReST. 
    Since all these interaction designs are commonly applied in the community, 
    and our theory describes their qualitative behaviors well despite strictly violating the assumption, 
    we believe our methods can shed more light on other practical methods with similar designs,
    like self-reward \citep{selfreward}, iterative-DPO \citep{xiong2023iterative}, etc.
    \item Identical $P_0(h)$ for agents in different generations. 
    Although this assumption makes it easier to derive \cref{prop:bayesian_il_converge}, 
    slightly relaxing it will not change the whole story: 
    we only require different agents to share a similar tendency towards a specific bias. 
    To verify this, we conduct several experiments when the agents in different generations are different LLMs 
    (e.g., \texttt{GPT3.5} plays with \texttt{Claude3} in \cref{append:exps_gpt} and \ref{append:exps_sec6}). 
    The phenomena claimed by the theory still hold.
    \item The Bayesian learning assumption, i.e., $P_{lm}(h)=P(h\mid \vd)\propto p(\vd \mid h) P_0(h)$.
    Although this assumption is necessary for drawing a parallel between iterated learning and the EM algorithm and hence getting a guarantee for the amplified bias,
    the practical in-weights learning (IWL for short) method usually does not strictly follow this assumption,
    because people usually early stop the training before the model perfectly learns all $\vd^t$.
    However, results in \cref{sec:applications_2} match our analysis well,
    which means the iterated IWL can also be depicted by iterated learning to some extent.
    That is because although there are plenty of finetuning methods with different targets or loss functions,
    their aims are consistent: 
    increasing the likelihood of $p(\vd_\text{train} \mid h)$ under instructions,
    which aligns with Bayesian targets well.
    Furthermore, we find the increased bias or decreased creativity during iterated finetuning has also been extensively mentioned in many related works \citep{touvron2023llama, xu2024perils},
    which also supports our analysis.
\end{itemize}

\subsection{Why we Start from Two ''Toyish'' Tasks}
The experimental settings in \cref{sec:exp_llm} and \ref{sec:applications} are relatively manual and toyish.
The main reason for us to start from them is that we want to \textit{directly observe} some quantitive numbers described by the theory,
which we believe would provide stronger support for the analysis.
For the explicit $h$ case, 
we chose the ACRE task because of its simple $\mathcal{H}$, 
making it possible to observe the distribution and entropy of all possible $h$.
We believe observing the model’s logits supports our theory more directly than merely observing the accuracy or other quantitative metrics.

For the implicit $h$ case, 
we chose the acronym task, 
which is a prototype of self-data-augmentation in self-instruct. 
We initially consider the conditional creative writing task (quite common in many related works),
where the model needs to write a passage (i.e., the list in our settings) 
based on several topic words (i.e., the acronym). 
However, constrained by the context length of LLMs, 
we can't generate more than 4 examples in one generation, 
which makes it hard to calculate the statistics of $\vd^{t}$. 
Remember the model will generate 20 extra $\vd^{t}$ based on 20 $\vd^{t-1}$ in our acronym experiments. 
In summary, although the experiments studied in our paper look artificial, they are reasonable approximations of real tasks.

Last, in a concurrent work \citet{xu2024perils}, 
the authors study practical applications like machine translation, 
creative writing, math reasoning, etc, in an iterative ICL setting. 
Their observations match our theoretical analysis quite well: 
bias is amplified generation by generation, and introducing external feedback can mitigate it. 
However, due to the complexity of the tasks they considered, 
they can only observe the average bias and the skew level using several conclusive quantitative metrics. 
Hence we believe that by combining our theoretical analysis, 
detailed observations on artificial examples, 
and the evidence from real applications in \citet{xu2024perils}, 
one can draw a good overview of how LLM would evolve in an iterated ICL setting.

\subsection{How our analysis brings benefits to practical algorithms}
Besides the method proposed in \cref{sec:applications_2},
where we manipulate the instructions prompt of the annotator LLM during the interaction phase,
our experiments and analysis also provide the following potential approaches to guide the model's evolution:
\begin{itemize}
    \item Select $\vd^0$ that makes more \textit{confident and correct} predictions on the target task. 
    Manually selecting good in-context examples is intuitive. 
    Our analysis, though, suggests taking the model’s confidence (i.e., the logits) into account, 
    because the theory claims that the likelihood and bias in prior are competing with each other during evolution.
    From \cref{fig:exp_entropy_convergence}, 
    we see the model evolves faster if the likelihood of $\vd^0 \mid h^*$ is weaker. 
    The results in \cref{fig:application_datapool} also provide similar insights: 
    the related biases are amplified slower when the number of easy samples in $\vd^0$ decreases.
    \item Designing a good interaction phase is important: 
    more accurate $\mathcal{H}_\text{eff}$ leads to better performance. 
    This can be supported by comparing self-refine and hypothesis-search in our paper. 
    The paper \citet{xu2024perils} also claims that external feedback with more accurate assessments or feedback from a larger model can reduce the amplified bias.
    \item Manipulating the instruction prompt: 
    in our analysis, 
    both $P_0(h)$ and $P_{lm}(h)$ are the model’s predictions conditioned on the instruction prompt $\vw$. 
    Hence adding preference in the task instruction (or changing the system prompt) during evolution could be an effective way of guiding the model’s evolution. 
    Our ACRE experiments show the feasibility of this: 
    remember we can introduce spurious correlation by adding one sentence to the instruction. 
    Hence it is also possible to guide the model’s evolution by feeding appropriate prompts during learning and sampling.
    \item Manipulating the temperature: 
    Bayesian-IL theory studies the evolution of the distribution, 
    so the temperature should also be an important factor for the evolution,
    as illustrated in \cref{fig:exp_entropy_convergence}. 
    We left the exploration between temperature and different phases in IL in the future.
\end{itemize}

\section{Proofs related to Bayesian Agents}
\label{append:proof_of_bayesian_il}

\begin{figure}[h]
    \centering
    \includegraphics[width=0.8\textwidth, trim={0, 10, 0, 0}, clip]{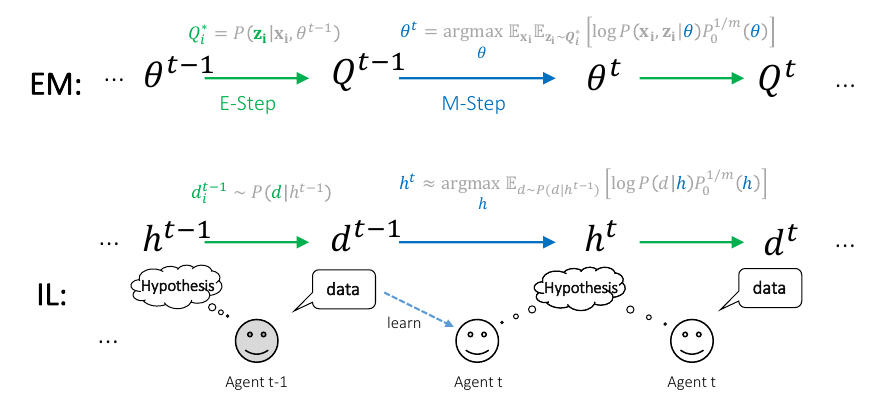}
    \caption{Illustrations of typical EM algorithm and an imitation-only iterated learning method.}
    \label{fig:app_iterated_learning_EM}
\end{figure}

\subsection{Recap the Proof of Expectation-Maximization Algorithm}

To get a clear picture of the asymptotic behavior of imitation-only iterated learning,
we first recap how a typical expectation-maximization (EM) algorithm converges when the target function is posterior distribution\footnote{The MLE (maximum likelihood estimation) version of the EM algorithm is more common in textbooks.}.
Consider a statistical model that generates a set of observable data samples $\{\vx_i\}_{i=1}^m$ and the corresponding hidden variables $\{\vz_i\}_{i=1}^m$.
The generating mechanism can be expressed as $P(\vx\mid\vz,\theta)$,
where $\theta$ is a set of unknown parameters determining this distribution.
To get a MAP(maximum a posterior) estimation of $\theta$,
we need to optimize the following target function:
\begin{align}
    \mathcal{L}(\theta) 
    &= \log P(\theta\mid\vx_1,\dots,\vx_m) \nonumber\\
    &= \log P_0(\theta) + \log P(\vx_1,\dots,\vx_m \mid \theta) - \log P(\vx_1,\dots,\vx_m) \nonumber\\
    &= \sum_{i=1}^m \frac{1}{m} \log P_0(\theta) + \sum_{i=1}^m \log P(\vx_i\mid\theta) - \log P(\vx_1,\dots,\vx_m) \nonumber\\
    &= \sum_{i=1}^m \log\left( P(\vx_i\mid\theta)P_0^{\frac{1}{m}}(\theta) \right) - \log P(\vx_1,\dots,\vx_m) \label{eq:L_of_EM}
\end{align}
where $P_0(\theta)$ is the prior distribution of parameters.
As the marginal distribution $P(\vx_i\mid\theta)$ is hard to calculate due to the existence of the hidden variable $\vz_i$,
our target function can then be expressed as
\begin{equation}
    \tilde{\mathcal{L}}(\theta) = \mathcal L(\theta) + \text{const} = \sum_{i=1}^m \log \left( \sum_{\vz_i} P(\vx_i, \vz_i\mid\theta)P_0^{\frac{1}{m}}(\theta) \right)
\end{equation}
where the term $\log P(\vx_1,\dots,\vx_m)$ is eliminated as it doesn't depend on $\theta$.
The target function above is still hard to tackle due to the summation inside the logarithmic function.
To solve this, we first introduce an auxiliary function $Q_i(\vz_i)$,
which is a probability distribution over $\vz_i$,
and reformulate the target as
\begin{equation}
    \tilde{\mathcal{L}}(\theta) = \sum_{i=1}^m \log \left( \sum_{\vz_i} Q_i(\vz_i)\frac{P(\vx_i, \vz_i\mid\theta)P_0^{\frac{1}{m}}(\theta)}{ Q_i(\vz_i)} \right)
    = \sum_{i=1}^m \log \left( \mathbb{E}_{\vz_i\sim Q_i} \left[\frac{P(\vx_i, \vz_i\mid\theta)P_0^{\frac{1}{m}}(\theta)}{ Q_i(\vz_i)}\right]\right). 
\end{equation}
Given the concavity of the logarithmic function,
we can use Jensen's inequality to get a lower bound of $\tilde{\mathcal{L}}(\theta)$:
\begin{equation}
    \tilde{\mathcal{L}}(\theta)\geq \mathcal{J}(\theta, Q) = \sum_{i=1}^m\mathbb{E}_{\vz_i\sim Q_i} \left[\log\frac{P(\vx_i, \vz_i\mid\theta)P_0^{\frac{1}{m}}(\theta)}{ Q_i(\vz_i)}\right]. \label{eq:L_of_EM_J}
\end{equation}
The EM algorithm then maximizes this lower bound by alternatively optimizing $\theta$ and $Q_i$ for several rounds.

In the \textbf{E-step},
as illustrated in \cref{fig:app_iterated_learning_EM},
we use the estimated $\theta^{t-1}$ in the previous round to find $Q_i^*=\text{argmax}_Q \mathcal{J}(\theta^{t-1}, Q)$.
Specifically, we need to push this lower bound to be tight by making the equality in \cref{eq:L_of_EM_J} hold.
Following the properties of Jensen's inequality,
the equality only holds when $\frac{P(\vx_i, \vz_i\mid\theta^{t-1})P_0^{\frac{1}{m}}(\theta)}{ Q_i(\vz_i)}$ is a constant.
Combining this requirement and the fact that $\sum_{\vz_i} Q_i(\vz_i)=1$,
we can calculate the optimal $Q_i^*(\vz_i)=P(\vz_i\mid\vx_i,\theta^{t-1})$,
which is the posterior distribution of $\vz_i$ given the observable data $\vx_i$ and the fixed parameters $\theta^{t-1}$.

In the \textbf{M-step},
we plug in the estimated $Q_i^*$ to $\mathcal{J}(\theta,Q)$ to get the target function as:
\begin{equation}
    \mathcal{J}(\theta;Q^*) = \sum_{i=1}^m\mathbb{E}_{\vz_i\sim Q^*_i} \left[\log\frac{P(\vx_i, \vz_i\mid\theta)P_0^{\frac{1}{m}}(\theta)}{ Q^*_i}\right]
    = \sum_{i=1}^m\mathbb{E}_{\vz_i\sim Q^*_i}\left[\log \left( P(\vx_i, \vz_i\mid\theta)P_0^{\frac{1}{m}}(\theta) \right) \right]-c,
    \label{eq:app_theta_target}
\end{equation}
where $c=\mathbb{E}_{\vz_i\sim Q^*_i}[Q^*_i]$ is a constant term and can be neglected while optimizing $\theta$.
In this step,
we can calculate $\theta^{t}=\text{argmax}_\theta \ \mathcal{J}(\theta;Q^*)$ using gradient descent or other parameter estimation methods.

In summary,
the E-step ensures a tight lower bound $\mathcal{J}(\theta,Q)$ and the M-step finds better $\theta$ to make it larger.
The two steps cooperate to ensure a series of estimations of $\theta$ for which $\mathcal{L}(\theta)$ is non-decreasing.
Finally,
the estimation of parameters will converge to the one that maximizes the posterior distribution,
i.e., $\mathbb{E}(\theta^*)=\text{argmax}_{\theta} \ P(\theta\mid \vx_1,\dots, \vx_m)$, if it is convex.

\subsection{Proof: Convergence Behavior of Bayesian Agents in Iterated Learning}
\label{append:proof_of_bayesian_il_body}

\bayesianilconverge*

\begin{proof}

The proof of this proposition can be divided into two steps.
In the first step,
we show that imitation-only iterated learning shares similar convergence behavior with a standard EM algorithm.
In the second step,
we show the ``selecting'' pressure introduced via the interaction phase doesn't break the necessary conditions of the convergence in the first step.
Merging these two steps leads to the proposition.

\textbf{Step 1: imitation-only iterated learning as a special EM}

Recall the imitation-only iterated learning illustrated in the bottom part in \cref{fig:app_iterated_learning_EM},
where the hypothesis held by the agent in the $(t-1)$-th generation is represented by $h^{t-1}$.
With this hypothesis,
the agent will generate $m$ data samples using $P(d\mid h^{t-1})$,
denoted $\vd^{t-1}\triangleq[d_1^{t-1},\dots,d_m^{t-1}]$.
In the $t$-th generation, 
a new agent will first update its posterior probability using $P(h|\vd^{t-1})$,
and then select $h^{t}$ by picking the one with the largest posterior.
As there are multiple data samples in $\vd^{t-1}$,
this process can be expressed as
\begin{align}
    \label{eq:app_h_target_def}
    h^{t} &= \argmax_h \log P(h\mid d_1^{t-1}, d_2^{t-1}, \dots ,d_m^{t-1})  \\
            &= \argmax_h \log\left( \frac{p(d_1^{t-1}, d_2^{t-1}, \dots ,d_m^{t-1}\mid h)P_0(h)}{P(d_1^{t-1}, d_2^{t-1}, \dots ,d_m^{t-1})} \right) \nonumber\\
            &= \argmax_h \log\left( P_0(h) \prod_{i=1}^m p(d_i^{t-1}\mid h) \right) \nonumber\\
            &= \argmax_h \frac{1}{m}\log P_0(h) + \frac{1}{m} \sum_{i=1}^m \log p(d_i^{t-1}\mid h) \nonumber\\
            &\approx \argmax_h \E_{d\sim p(d|h^{t-1})} [\log P_0^{\frac{1}{m}}(h)] + \E_{d\sim p(d|h^{t-1})} [\log p(d\mid h)] \nonumber\\
            &= \argmax_h \E_{d\sim p(d\mid h^{t-1})} [\log p(d\mid h)P_0^{\frac{1}{m}}(h)].\label{eq:app_h_target}
\end{align}

Based on the analysis above,
we notice that the imitation-only iterated learning and the EM algorithm are almost identical:
by replacing $\theta$ and $\vz$ to $h$ and $\vd$,
and by removing the random variable $\vx$\footnote{It is unusual to apply EM with no observable data, but removing it doesn't violate any assumptions in the derivation of EM.},
we can also have similar theoretical guarantees for the imitation-only iterated learning algorithm.

To prove this,
we can first verify the equivalence between the imitation phase and an M-step.
By comparing the target functions when calculating hidden variables ($h$ and $\theta$) in these two algorithms, i.e.,  \cref{eq:app_theta_target} and (\ref{eq:app_h_target}),
we can find two major differences.
First, the expectation of observable samples (i.e., $\mathbb{E}_{\vx_i}$) disappears in \cref{eq:app_h_target},
as we assume there are no ``observations'' in iterated learning.
We can also introduce a dummy variable named $\vx$ to the iterated learning process,
and find that the existence of $\vx$ doesn't influence the aforementioned calculation at all.
Second, in IL,
we can only approximate $\mathbb{E}_{d_i^{t-1}}[\cdot]$ by sampling $d$ from $p(d\mid h^{t-1})$,
while in EM,
the posterior distribution $P(\vz_i\mid \vx_i, \theta^t)$ is usually analytically calculated.
Of this discrepancy,
if our $\vd$ is a good approximation of $p(d\mid h^{t-1})$,
the imitation phase in IL is equivalent to an M-step in EM.

We then verify whether the transmission phase is a good approximation of an E-step.
The first thing to check is the tightness of the lower bound generated via Jensen's inequality,
which guarantees the non-decreasing update of the target function across multiple generations.
we can first assume the target function of the whole iterated learning process is $\mathcal{L}(h)=\log P_0(h)$,
and derives its lower bound $\mathcal{J}(h;Q)$ following a similar procedure in EM:
\begin{align}
    \mathcal{L}(h) &= \log P_0(h) \nonumber\\
    &= m\log P_0^{\frac{1}{m}}(h) \nonumber\\
    &= m\log\left( \sum_{d_i} p(d_i\mid h) P_0^{\frac{1}{m}}(h) \right)\nonumber\\
    &= m\log\left( \sum_{d_i} Q_i(d_i)\frac{p(d_i\mid h) P_0^{\frac{1}{m}}(h)}{Q_i(d_i)}\right) \nonumber\\
    &= m\log \mathbb{E}_{d_i\sim Q_i}\left[ \frac{p(d_i\mid h) P_0^{\frac{1}{m}}(h)}{Q_i(d_i)} \right] \nonumber\\
    &\geq \mathbb{E}_{d_i\sim Q_i}\log \left[ \frac{p(d_i\mid h) P_0^{\frac{1}{m}}(h)}{Q_i(d_i)} \right] \triangleq \mathcal{J}(h,Q)
    \label{eq:app_Lh_def}
\end{align}
The equality of Jensen's inequality holds when $\frac{p(d_i\mid h) P_0^{1/m}(h)}{Q_i(d_i)}$ is a constant.
Using the fact that $\sum_{d_i}Q_i(d_i)=1$,
we can have the optimal $Q_i^*=p(d_i\mid h)$.
In summary,
as we sample each $d_i^{t-1}\sim p(d\mid h^{t-1})$,
the transmission phase in IL is equivalent to an E-step in EM.

Another interesting parameter to discuss is $m$,
i.e., the number of data samples generated by an agent in each generation.
The choice of $m$ determines how well the sampled $\vd$ can represent the ground truth $p(d\mid h)$.
For large enough $m$,
we can prove the convergence guarantee using the above procedure.
When $m=1$,
the standard EM algorithm becomes a stochastic EM approximation \cite{gilks1995markov}.
The authors of \citet{nielsen2000stochastic} proved that in stochastic EM,
$\theta$ in different generations form a homogeneous Markov chain whose stationary distribution over hypotheses is approximately centered on the maximum-likelihood solution.
In other words,
when $t>T$ for sufficiently large $T$,
$\mathbb{E}[\theta^t]$ optimizes $\mathbb{E}_{\vx}[P(\theta\mid\vx_1,\dots,\vx_m)]$,
and similarly,
$\mathbb{E}[h^t]$ is the optimizer of $P_0(h)$.
In other words,
the dominating hypothesis in imitation-only iterated learning converges to the one with the highest prior,
which is equivalent to the large $m$ case.

Although $m=1$ doesn't influence the converged estimation of $h$,
a too small $m$ will make the variance of estimation large,
and hence impede the converging speed of $h$.
Then should we choose $m$ as large as possible?
The answer is still no: too large $m$ will also impede the converging speed.
We can get some intuition by observing \cref{eq:app_h_target},
where $P_0^{1/m}(h)$ determines the effect of the prior when selecting optimal $h$ in each generation.
If $m$ is too large,
this distribution would be very flat and the preference encoded in the prior cannot influence the choice of $h$ in this generation much --
the likelihood term $P(d\mid h)$ will dominate.\footnote{A similar trend also exists in Bernstein-von Mises theorem \citep[see e.g.][]{van2000asymptotic}, which claims the posterior $p(\theta\mid x_1,\dots,x_n)=\mathcal{N}(\theta_0,n^{-1}I(\theta_0)^{-1})$, for $n\rightarrow\infty$.
In other words, $P(h|d_1,\dots,d_m)$ would become peakier as $m$ increases, and hence $h^t$ will be closer to $h^{t-1}$ after the imitation phase.}
Hence the resulting $h^t$ would be quite close to $h^{t-1}$,
which means the evolution of belief on $h$ would be slow.

Actually, the choice of $m$ is usually considered as the ``bottleneck'' parameter in different iterated learning algorithms.
Almost all the related studies point out that the bottleneck should not be too wide or too tight (like experiments in \citet{kirby2015compression} and \citet{nil}).
Our Bayesian analysis provides a theoretical explanation of the effect.

\textbf{Step 2: the influence of introducing $\mathcal{H}_\text{eff}$.}
Let us first check the imitation phase,
i.e., the E-step where $h^t$ is calculated in \cref{eq:app_h_target_def}.
Assume we have a perfect interaction phase that can rule our all $h\notin \mathcal{H}_\text{eff}$,
then the target function is:

\begin{align}
    \label{eq:app_h_target_def_interaction}
    h^{t} &= \argmax_{h\in\mathcal{H}_\text{eff}} \log P(h\mid d_1^{t-1}, d_2^{t-1}, \dots ,d_m^{t-1})  \\\nonumber
            &= \argmax_{h} \log P(h\mid d_1^{t-1}, d_2^{t-1}, \dots ,d_m^{t-1})\cdot\mathds{1}(h\in\mathcal{H}_\text{eff}) \\\nonumber
            &= \argmax_h \E_{d\sim p(d\mid h^{t-1})} \left[\log p(d\mid h)\cdot\left( P_0^{\frac{1}{m}}(h)\cdot\mathds{1}(h\in\mathcal{H}_\text{eff})\right) \right] \\\nonumber
            &\triangleq \argmax_h \E_{d\sim p(d\mid h^{t-1})} \left[\log p(d\mid h)\cdot \tilde{P}_0^{\frac{1}{m}}(h) \right],
\end{align}
where we define a ``regularized'' prior as $\tilde{P}_0(h)\triangleq c\cdot P_0(h)\cdot\mathds{1}(h\in\mathcal{H}_\text{eff})$.
Then, by substituting this prior back to $\mathcal{L}(h)$ defined in \cref{eq:app_Lh_def},
we find the optimal $Q_i^*=p(d_i\mid h)$ still holds.
In other words,
as long as we same $d_i^{t-1}\sim p(d\mid h^{t-1})$, where $h^{t-1}\in\mathcal{H}_\text{eff}$,
all the conditions required for this proposition still hold.
Furthermore,
this proof also provides us an insight that adding constraints on $h\in\mathcal{H}_\text{eff}$ is required in both imitation and transmission phases.
Hence having a powerful ``data-filter'' or ``data-ranking'' design for the transmission phase would also make the evolution more robust,
like those applied in RAFT \citep{dong2023raft} and ReST \citep{rest}.
\end{proof}

\subsection{Proof: LLM as a Bayesian Agent}
\label{append:llm_bayesian_proof}
\iclisil*

\begin{proof}
    The proof of this proposition can be divided into two steps.
    In the first part,
    we de-marginalize the posterior predictive distribution on a hidden variable $h$,
    and then show that the model automatically ``selects'' a hypothesis $h^{t*}$ that generates the prompting examples.
    In the second part,
    we show that when the examples in $\vd^{t-1}$ are generated by another LLM with the same prior belief over $h$,
    the MAP (maximize a posterior) estimation of $h$ can approximate $h^{t*}$ well.

    \textbf{Step 1:} In our paper, 
    we assume the query and answer sequences in each example, i.e., $d_i=(x_i,y_i)$,
    are controlled by the hidden hypothesis $h$,
    which plays a similar role to the ``concept'' parameter $\theta$ mentioned in \citet{xie2022an}\footnote{As we only need the ``selection'' mechanism mentioned in this paper, the HMM (Hidden Markov Model) assumption is not required in our setting.}.
    Then the posterior predictive distribution can be decomposed as:
    \begin{align}
        P(d\mid \vd^{t-1}) 
        &= \int_h p(d\mid \vd^{t-1}, h) P(h\mid \vd^{t-1}) \ \dd h \nonumber\\
        &= \int_h p(d\mid h) P(h\mid \vd^{t-1}) \ \dd h \nonumber\\
        &\propto  \int_h p(d\mid h) p(\vd^{t-1}\mid h) P_0(h) \ \dd h \nonumber\\
        &\propto \int_h p(d\mid h) \frac{p(\vd^{t-1}\mid h)}{p(\vd^{t-1}\mid h^{t*})} P_0(h) \ \dd h \nonumber\\
        &= \int_h p(d\mid h) \exp (n \cdot r_n(h)) P_0(h) \ \dd h, \label{eq:app_Pd_d}
    \end{align}
    
    where $r_n(h)\triangleq \frac{1}{n}\log\frac{p(\vd^{t-1}\mid h)}{p(\vd^{t-1}\mid h^{t*})}$.
    In \cref{eq:app_Pd_d},
    the second line follows the Markov property of hypothesis and data samples,
    the third line follows the Bayesian rule and drops a constant term,
    the fourth line is generated by dividing a constant $p(\vd^{t-1}\mid h^{t*})$,
    where $h^{t*}$ is a hypothesis that generates $\vd^{t-1}$.
    Now we see the $r_n(h)$ has almost the same form as $r_n(\theta)$ in \citet{xie2022an}.
    By reusing the derivation in that paper (mainly in Section 3.2 and the proof of its Theorem 1),
    we can conclude that $\exp (n \cdot r_n(h))\rightarrow 0$ for any $h\neq h^{t*}$ and $\exp (n \cdot r_n(h^{t*}))\rightarrow 1$.
    Hence \cref{eq:app_Pd_d} becomes:
    \begin{equation}
        P(d\mid \vd^{t-1}) = P(d\mid h^{t*}).
    \end{equation}

    \textbf{Step 2:} In a general in-context learning setting,
    the prompting examples $\vd^{t-1}$ are usually sampled from an unknown distribution $P_{\text{prompt}}$.
    For example, by analyzing different $x_{\text{test}}$,
    the researchers can manually design effective prompting examples sharing similar chain-of-thought structures with the target question.
    Obviously, such $P_{\text{prompt}}$ is impossible to parameterize or analyze accurately.
    To approximate it, \citet{xie2022an} first assumes that both the prompting examples and the pre-training corpus are natural languages,
    and a well-trained LLM can approximate this ``natural language distribution'' well.
    Then, $P_{\text{prompt}}$ can be well approximated by $P(\cdot\mid \theta^*)$ under some $\theta^*$,
    which is the prompt concept in that paper.

    In our settings,
    as we assume the prompting examples $\vd^{t-1}$ are generated by another agent-B sharing the \textit{same prior},
    then the $h^{t*}$ triggered by feeding $\vd^{t-1}$ to agent-A would be exactly the same as that feeding that to agent-B,
    i.e., 
    \begin{equation}
        h^{t*} = \argmax_h P_{lmw}(h\mid \vd^{t-1}),
    \end{equation}   
    where $P_{lmw}$ is model's belief after receiving the common instruction $\vw$.
    
Combining these two steps, we can decompose the sampling procedure $d^t\sim P(d\mid \vd^{t-1})$ into two parts:
first, inherently select $h^{t*}$ based on observations generated by another agent in the previous generation;
then sample new $d$ conditioned on this $h^{t*}$, 
which matches the Bayesian-IL procedure discussed in this paper.
\end{proof}

\section{Experiments of Iterated Learning on Different Domains}
\label{append:exps_of_il}

To provide a panorama of iterated learning and its applications in different fields,
this appendix will first give an intuitive explanation using some lab experimental results.
Then, experiments on the emergence of compositional language among Bayesian agents are introduced to verify all the theoretical hypotheses.

\subsection{Iterated Learning in Lab Experiments using Lewis Language Game}

\begin{figure}[h]
    \begin{center}
    \centerline{\includegraphics[width=0.9\textwidth]{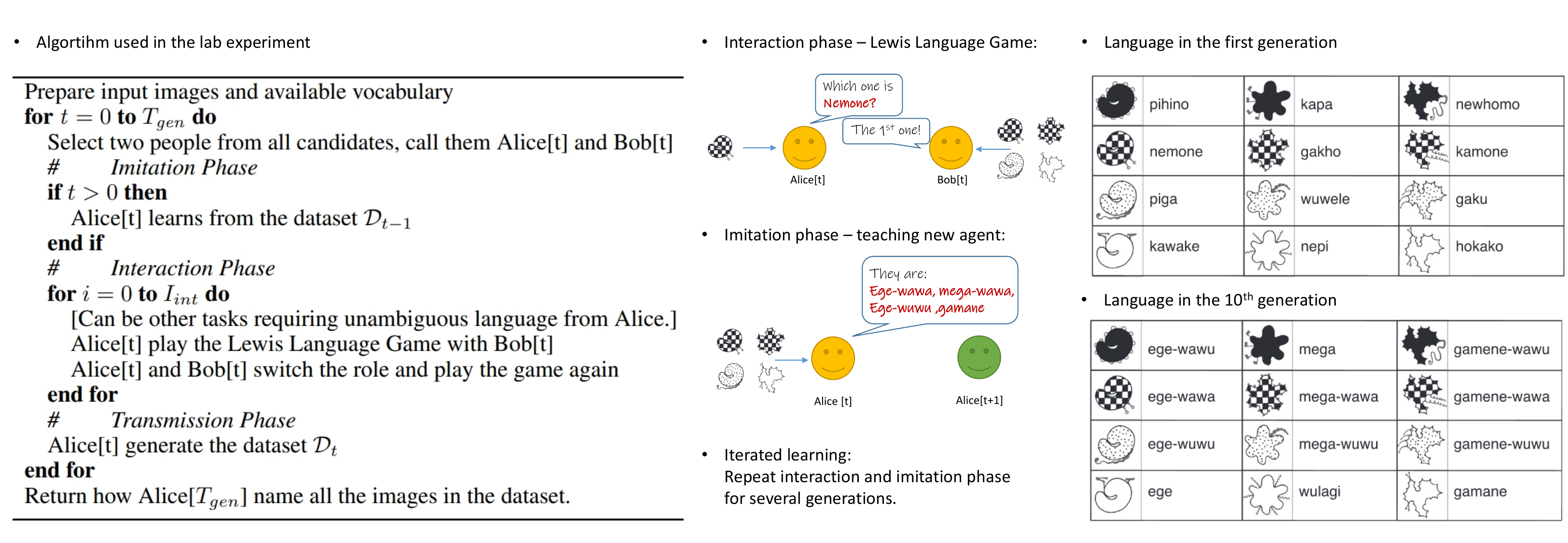}}
    \caption{The lab experiments (algorithm, settings, and results) conducted in \cite{kirby2015compression}.}
    \label{fig:app_iterated_learning}
    \end{center}
\end{figure}

As denoted in \cite{kirby2008cumulative},
iterated learning is a process where one individual learns by observing the output of another individual, who learned in the same way.
In this multiple-generation learning procedure,
the shared language (i.e., $h$) among learning agents will gradually become more systematic under the compressibility pressure (imposed during imitation, embodied in $P_0(h)$) and expressivity pressure (imposed during interaction, requiring $h\in\mathcal{H}_\text{eff}$).
To simulate this process,
authors of \cite{kirby2015compression} design a two-phases learning procedure illustrated in Figure~\ref{fig:app_iterated_learning}.
In the interaction phase,
the speaker (Alice) and listener (Bob) must cooperate to accomplish a Leiws referential game (see the middle panel in Figure~\ref{fig:app_iterated_learning}).
Specifically,
Alice will first create a name for the given objects and talk that to Bob.
After receiving this message, Bob needs to select the correct object shown to Alice among some candidates.
If correct, both of them are rewarded. This phase terminates when they can achieve a high enough success rate.
To succeed in this game,
the shared language should be expressive enough to avoid any ambiguities --
we expect the language to be a bijection.
Then, we select another new naive candidate (Alice[t+1]) and let it learn the naming system created by Alice[t] and Bob[t].
In this phase,
those highly structural mappings should be easier for a human to remember,
which is how the compressibility pressure is imposed.
After that, Alice[t+1] will play the same game with another Bob[t+1] and the interaction phase starts again.
We provide the languages generated by Alice[0] and Alice[10] in the right panel in Figure~\ref{fig:app_iterated_learning}:
it is clear that an interesting structure emerges in the language generated by Alice[10].
There are also plenty of similar lab experiments that support the ``two pressures'' and cultural evolution hypothesis using IL-like training methods,
e.g., \cite{fay2010interactive, ferdinand2019cognitive, motamedi2019evolving, motamedi2022improvisation}.
Although these methods have different types of input, game designs, learning procedures, vocabularies (like gesture language), etc.,
the conclusion of them is quite consistent:
compressibility and expressivity pressures are crucial for the emergence of systematic mappings,
iteratively learning and interacting can amplify these pressures a lot,
which matches the Bayesian explanations well.

\subsection{Iterated Learning of Bayesian Agents (re-implementation of results in \citet{kirby2015compression})}
\label{append:exps_of_il:bayesian256}
To verify that the emergence of systematic mappings in iterated learning is not an accident,
authors of \cite{beppu2009iterated} provide a guarantee by analyzing the behavior of Bayesian agents.
There are also plenty of related works in cognitive science, like \cite{bonawitz2014probabilistic, ferdinand2014regularization, navarro2018extremists, perfors2011tutorial, reali2009evolution}, discussing the influence of and theories behind iterated learning and Bayesian analysis.

\begin{figure}[h]
    \begin{center}
    \centerline{\includegraphics[width=0.95\textwidth, trim={50, 10, 50, 250}, clip]{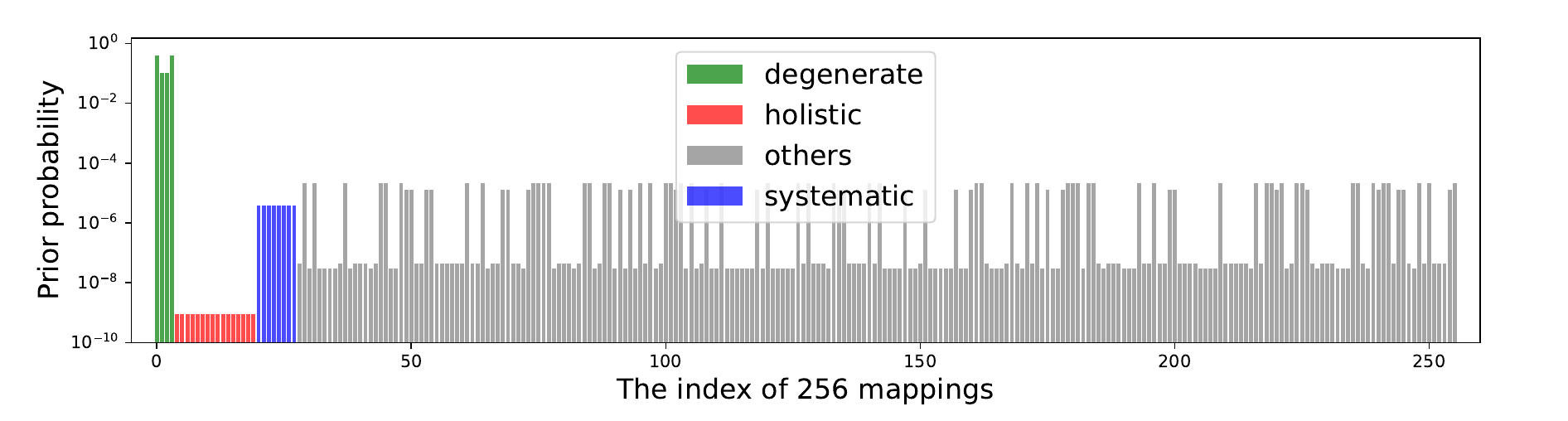}}
    \caption{The prior probability of all possible $h\in\mathcal{H}$.
            The systematic mappings are sandwiched between degenerate and holistic mappings,
            which means $P_0(h_{degen})>P_0(h_{sys})>P_0(h_{holi})$.
            Some mappings in the ``other'' group also have relatively large prior,
            because they contain degenerate components (e.g., mapping two or three objects to the same message).}
    \label{fig:app_topsim_prior}
    \end{center}
\end{figure}

To give the readers a better understanding of how iterated learning works,
we re-implement the Bayesian experiments mentioned in \cite{kirby2015compression}.
Consider the following toy example,
where we have four different input objects: $\mathcal{X}=\{ \texttt{blue circle, blue box, red circle, red box} \}$,
and four possible names: $\mathcal{Y}=\{\texttt{00,01,10,11}\}$.
The hypothesis $h$ is defined as $h\in\mathcal{H}:\mathcal{X}\rightarrow \mathcal{Y}$.
In this example, we have $|\mathcal{H}|=256$,
which means $P(h)$ can be parameterized by a categorical distribution with 256 dimensions.
In this analysis,
we assume the prior distribution of a mapping is negatively correlated with its \textit{coding length} $\alpha$,
i.e., $P(h;\alpha, c)\propto 2^{-\frac{\alpha}{c}}$,
where $c$ is a normalizing constant to make sure the prior distribution is not too peaky.
Usually, the easier-to-learn mappings (i.e., more systematical ones) have higher prior.
In Table~\ref{tab:app_coding_length},
we demonstrate how to calculate the coding length for the three typical mappings.
Note that the mapping that has the highest prior is a degenerate mapping,
where $\alpha=18$ and $P(h)\approx 0.6$.
The $P_0(h)$ for all possible mappings are demonstrated in Figure~\ref{fig:app_topsim_prior}.

\begin{table}[h]
    \vskip 0in
    \caption{An example of coding the mappings, where $\alpha$ is how many characters 
             (including space and unique symbol, e.g., $\rightarrow$ and $:$) are used to express the grammar.}
    \centering\begin{tabular}{cccccccccccc}
    \hline
      \multicolumn{4}{c}{\begin{tabular}[c]{@{}c@{}}A systematic mapping\\ $\alpha=43$\end{tabular}} & 
      \multicolumn{4}{c}{\begin{tabular}[c]{@{}c@{}}A holistic mapping\\ $\alpha=56$\end{tabular}} & 
      \multicolumn{4}{c}{\begin{tabular}[c]{@{}c@{}}A degenerate mapping\\ $\alpha=18$\end{tabular}} 
      \\ \hline
    S   &   & $\rightarrow$ & z2, z1 &  &    &                  &             &   & & & \\
    z2: & 0 & $\rightarrow$ & blue   &  & S: & 00 $\rightarrow$ & blue circle &     &                  &  \\
    z2: & 1 & $\rightarrow$ & red    &  & S: & 01 $\rightarrow$ & red circle  &  & S: & 00 $\rightarrow$ & Everything\\
    z1: & 0 & $\rightarrow$ & circle &  & S: & 10 $\rightarrow$ & red box     &     &                  &  \\
    z1: & 1 & $\rightarrow$ & box    &  & S: & 11 $\rightarrow$ & blue box    &     &                  &  \\ \cline{1-12}
    \end{tabular}
    \label{tab:app_coding_length}
\end{table}

In this experiment,
the knowledge of an agent is encoded in its posterior distribution, i.e., $P_{lm}(h)$.
We will observe how this distribution evolves when the agents conduct iterated learning.
Recall the sampling behaviors discussed in \cref{sec:bayesian_il}.
To get a data sample $d=(x,y)$,
we first randomly sample $x\in\mathcal{X}$ from a uniform distribution and then sample $y$ based on the given $x$,

\begin{equation}
    d\sim P_{lm}(x,y)\propto P_{lm}(y\mid x) \propto p(y\mid h, x)\cdot P_{lm}(h).
    \label{eq:app_agent_sampling}
\end{equation}

The likelihood $p(y\mid h, x)$ is defined as:
\begin{equation}
    p(y\mid h, x) = \left\{
        \begin{aligned}
            1-\epsilon   \ \ &  \qquad \text{if } y \text{ is mapped to } x \text{ in } h\\
            \frac{\epsilon}{|\mathcal{Y}|-1}  & \qquad \text{otherwise,}
        \end{aligned}
    \right.
    \label{eq:app_likelihood}
\end{equation}
where $\epsilon$ is a small positive value describing the systematic error during communication.

For the learning behavior,
the agent will update the posterior based on the received data samples $\vd=(x_i,y_i)_{i=1}^N$:
\begin{equation}
    P_{lm}(h)=P(h\mid\vd)\propto p(\vd\mid h)\cdot P_0(h) \propto P_0(h)\cdot \prod_{i=1}^N p(y_i\mid h, x_i).
    \label{eq:app_agent_learning}
\end{equation}

Now, with the definition of learning and sampling for these Bayesian agents,
we can describe how they conduct IL:

\begin{itemize}
    \item Initialization: at the beginning of the $t$-th generation, a new agent is initialized by $P_0(h)$
    \item Imitation: agent-$t$ learns from $\vd^{t-1}$, which is generated by agent in the previous generation following \cref{eq:app_agent_learning}
    \item Interaction: to impose expressivity pressure, we let agent-$t$ (Alice) play a communication game in this phase.
            Specifically, we first create another agent Bob by copying $P_{lm}(h)$ from Alice.
            Then, Alice samples a data pair $d=(x,y)$ on a randomly chosen $x$ and sends it to Bob.
            Bob will estimate the object based on $y$.
            If the estimated $x'=x$, the game succeeds and data pair $(x,y)$ is added to a buffer named $\vd_\text{comm}$.
            After several rounds,
            Alice updates its knowledge by learning from $\vd_\text{comm}$.
            Note that in this phase,
            the pressure of $h\in\mathcal{H}_\text{eff}$ is induced implicitly:
            for the ambiguous $h$, where multiple $x$ are mapped to the same $y$,
            Bob's reconstruction $x'$ might not equal $x$ with high probability.
            Hence $\vd_\text{comm}$ will finally dominated by the samples generated by those $h\in\mathcal{H}_\text{eff}$.
    \item Transmission: after the interaction phase,
            Alice will generate multiple samples $\vd^{t}$ for the next generation.
\end{itemize}

\begin{figure}[h]
\vskip -0.05in
    \begin{center}
    \centerline{\includegraphics[width=0.85\textwidth, trim={130, 0, 140, 0}, clip]{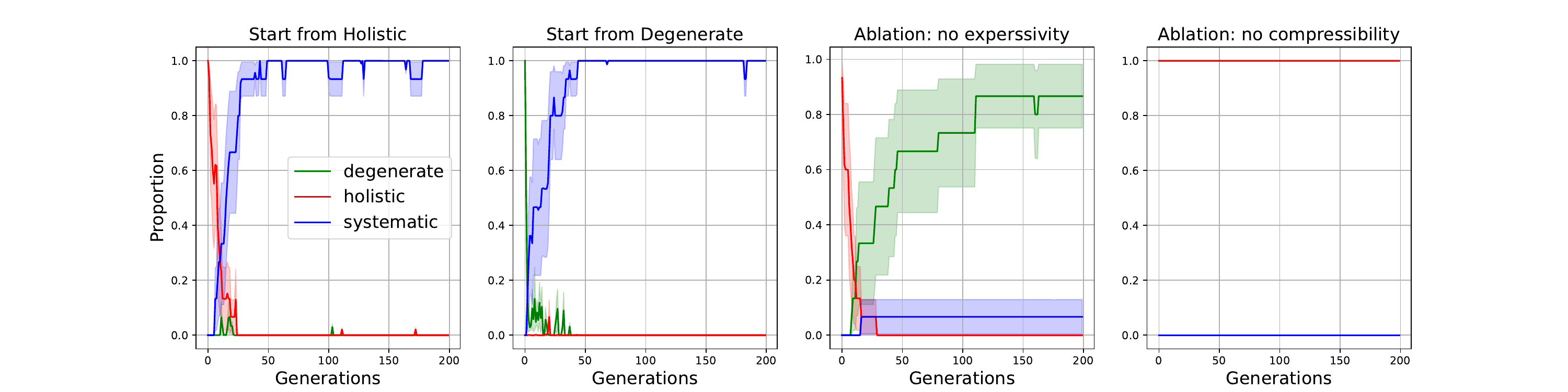}}
    \caption{Ratio of three different types of mappings during iterated learning (curves are the average of 15 different runs, shadow region is the variance).
            Left to right: 1.) $\vd^0$ is a holistic mapping;
            2.) $\vd^0$ is a degenerate mapping;
            3.) Starting from a holistic $\vd^0$, but no longer conduct an interaction phase during training.
                Hence the degenerate language, which has the highest prior, will gradually dominate;
            4.) Ablating the compressibility pressure by using a uniform prior distribution.}
    \label{fig:app_ratio_language}
    \end{center}
\vskip -0.2in
\end{figure}

\begin{figure}[h]
\vskip 0 in
    \begin{center}
    \centerline{\includegraphics[width=0.85\textwidth, trim={50, 50, 50, 50}, clip]{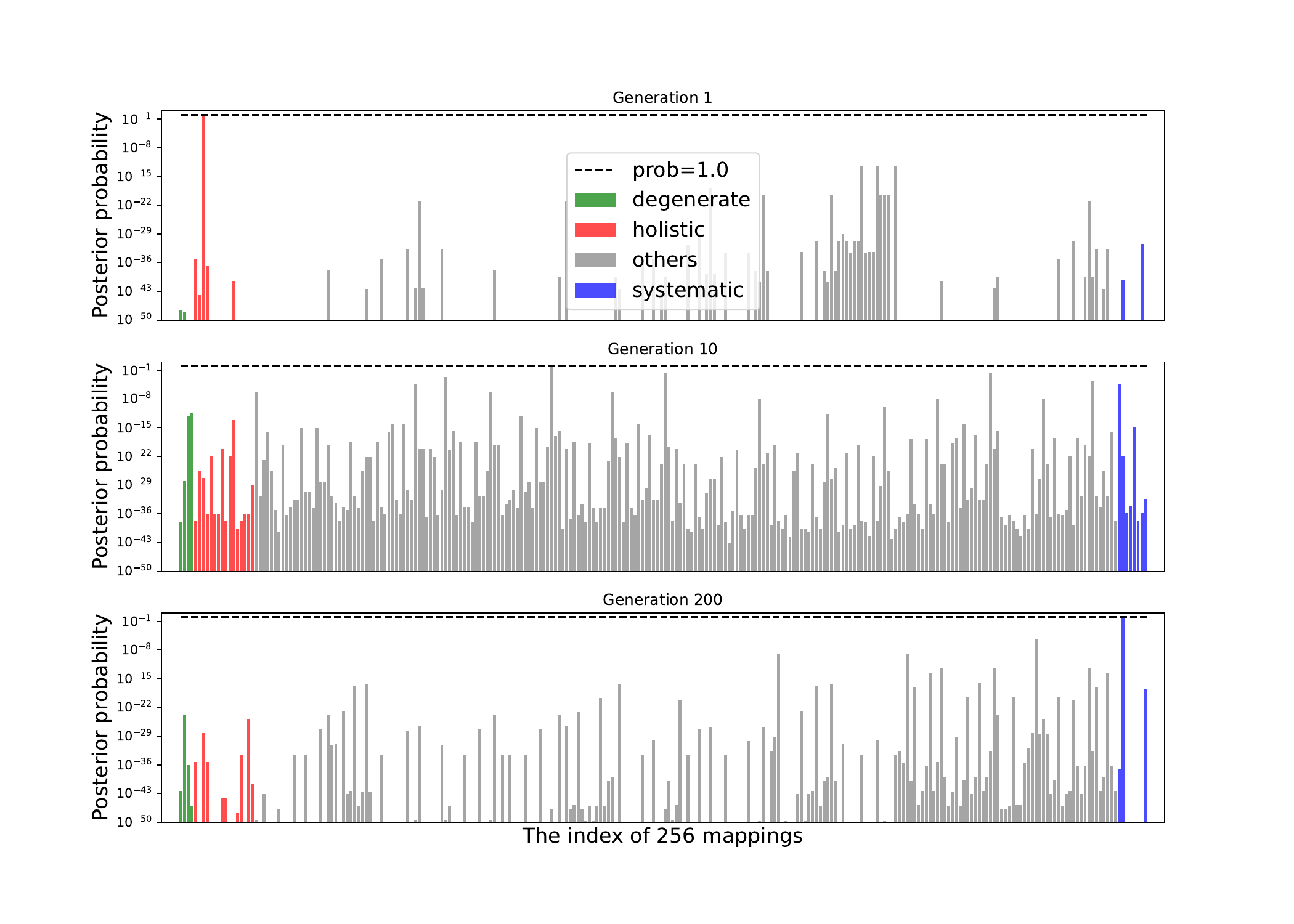}}
    \caption{The posterior probabilities of all at the end of different generations.}
    \label{fig:app_posterior}
    \end{center}
\vskip -0.2in
\end{figure}

In the above procedure,
the compressibility pressure is embodied in the prior distribution where more reusable principles lead to a higher probability,
which aligns with the simplicity bias in the human cognition system.
The expressivity pressure is imposed in the communication game
because $\vd_\text{comm}$ only contains the unambiguous mappings.
Under this setting,
we can calculate the weighted proportion 
(i.e., the summation of the posteriors) 
of different types of languages and observe how they evolve during iterated learning,
as illustrated in Figure~\ref{fig:app_ratio_language}.
It is clear that the systematic mappings gradually dominate as the learning progresses.

To further verify our theory,
we consider a compressibility-only case by removing the interaction phase,
and an expressivity-only case using a uniform prior $P(l)=1/256$.
The results in Figure~\ref{fig:app_ratio_language} match the theory quite well:

\begin{itemize}
    \item In an imitation-only iterated learning case, i.e., the third panel in \cref{fig:app_ratio_language},
          $h^T$ converges to a degenerate mapping, which has the highest prior as illustrated in \cref{fig:app_topsim_prior};
    \item Introducing the interaction phase will rule out those ambiguous mappings (i.e., those $h\notin\mathcal{H}_\text{eff}$), 
          and hence $h^T$ converges to systematic mapping, which has the highest prior among all $h\in\mathcal{H}_\text{eff}$;
    \item By comparing the first and second panels in \cref{fig:app_ratio_language},
          $h^T$ always converges to systematic mapping no matter $\vd^0$ is holistic or degenerate.
\end{itemize}

Furthermore, we can directly observe the dynamics of $P_{lm}(h)$ from Figure~\ref{fig:app_posterior},
which provides a more detailed illustration of how the posterior of all mappings changes during training.
In the first generation, we see the dominant mapping is a holistic one,
which is our $\vd^0$. 
Then gradually, under the two pressures, 
the posterior of systematic mappings gradually increases and finally dominates.

\section{More on GPT-based ACRE Experiments}
\label{append:exps_gpt}

\begin{figure}[h]
\vskip -0.05in
    \begin{center}
    \centerline{\includegraphics[width=0.8\textwidth]{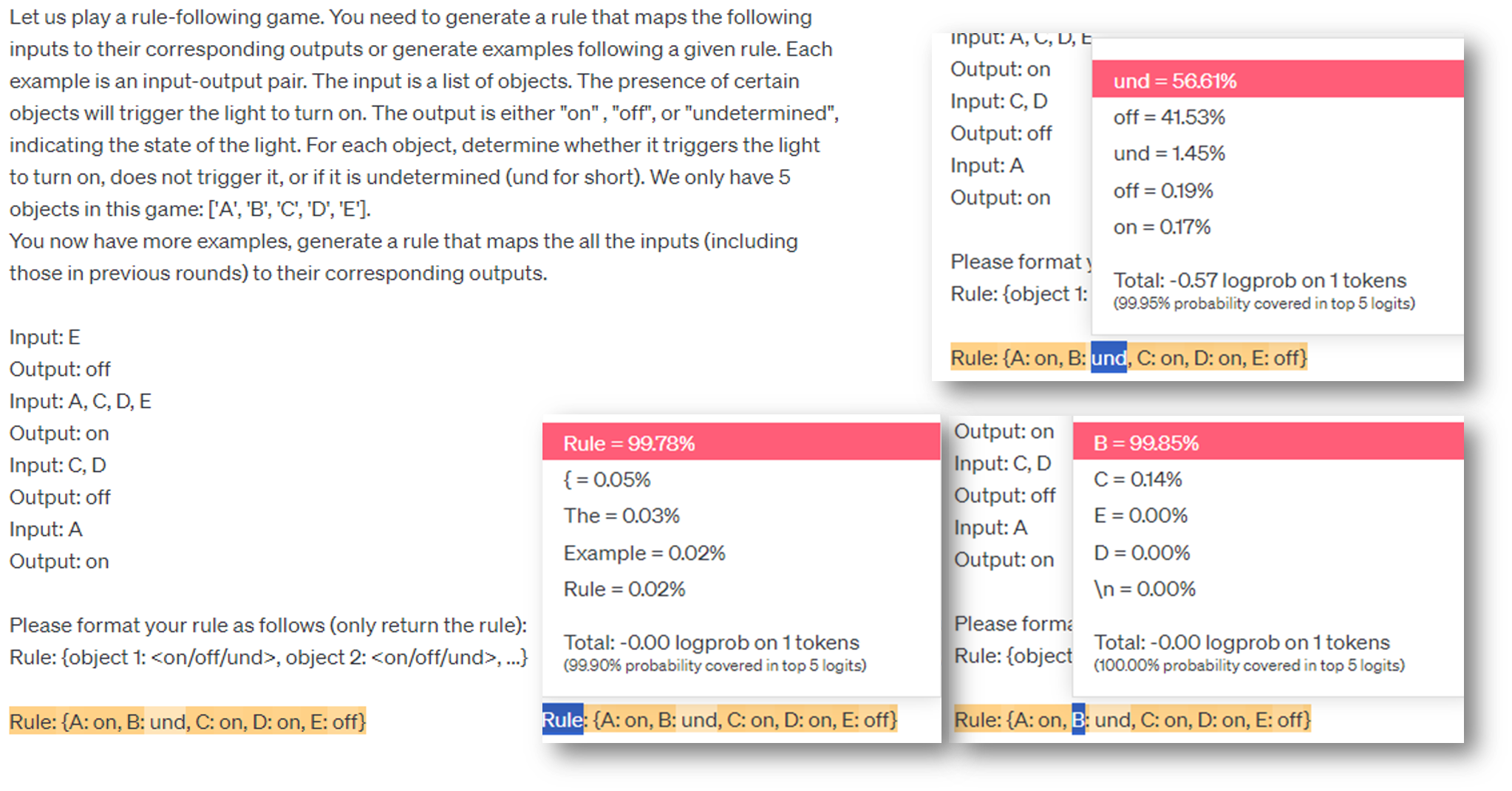}}
    \caption{How GPT provides the rule following the given format,
             which makes it possible to calculate $P_{lmw}(h)$ for all $h$.}
    \label{fig:app_gpt_cal_ph}
    \end{center}
\vskip -0.2in
\end{figure}

\subsection{How to Calculate the Model's Posterior on All Hypotheses (\cref{fig:app_gpt_cal_ph})}
\label{append:exps_gpt:calculate_ph}

Thanks to the instruction-following ability,
the GPT can always provide responses following the given format,
as illustrated in \cref{fig:app_gpt_cal_ph}.
The experiments demonstrated in this part come from OpenAI's playground.
The model we use is \texttt{gpt-3.5-turbo-instruct}.
The temperature is 0.1 and the probability feedback is enabled.
We let the model return probabilities of the top 5 candidate tokens for each token in the response,
as illustrated by the three sub-panels in the figure.
Then, the posterior of specific $h$ can be calculated by multiplying the probability of all tokens with corresponding values.
For example, $P_{lmw}(h=\texttt{\{A:on, B:und, C:off, D:...\}})$ can be calculated by $P(r_5=\texttt{on})\cdot P(r_9=\texttt{und})\cdot P(r_{13}=\texttt{off})\cdots$,
where $r_{5,9,13,...}$ are the tokens denoting the corresponding values of \texttt{A, B, C}.
To further show the feasibility of this approach,
we conduct the following two verifications.
First,
we calculate $\prod_{i\in\mathcal{I}_\text{format}}P(r_i)$ and find that this value is always close to one 
($\mathcal{I}_\text{format}$ denotes the indexes of those format-related tokens, e.g., \texttt{Rule}, \texttt{\{}, \texttt{:}, \texttt{A}, \texttt{B}, \texttt{C}, etc).
This means the model reliably follows the instructions when generating responses.
Second,
we calculate $P_{lmw}(h)$ for all possible 243 different $h$ and verified that $\sum_{h\in\mathcal{H}}P_{lmw}(h)$ is always close to one.

\subsection{How to Control the Bias in the Prior}
\label{append:exps_gpt:prior_control}

\begin{figure}[h]
\vskip -0.05in
    \begin{center}
    \centerline{\includegraphics[width=0.8\textwidth]{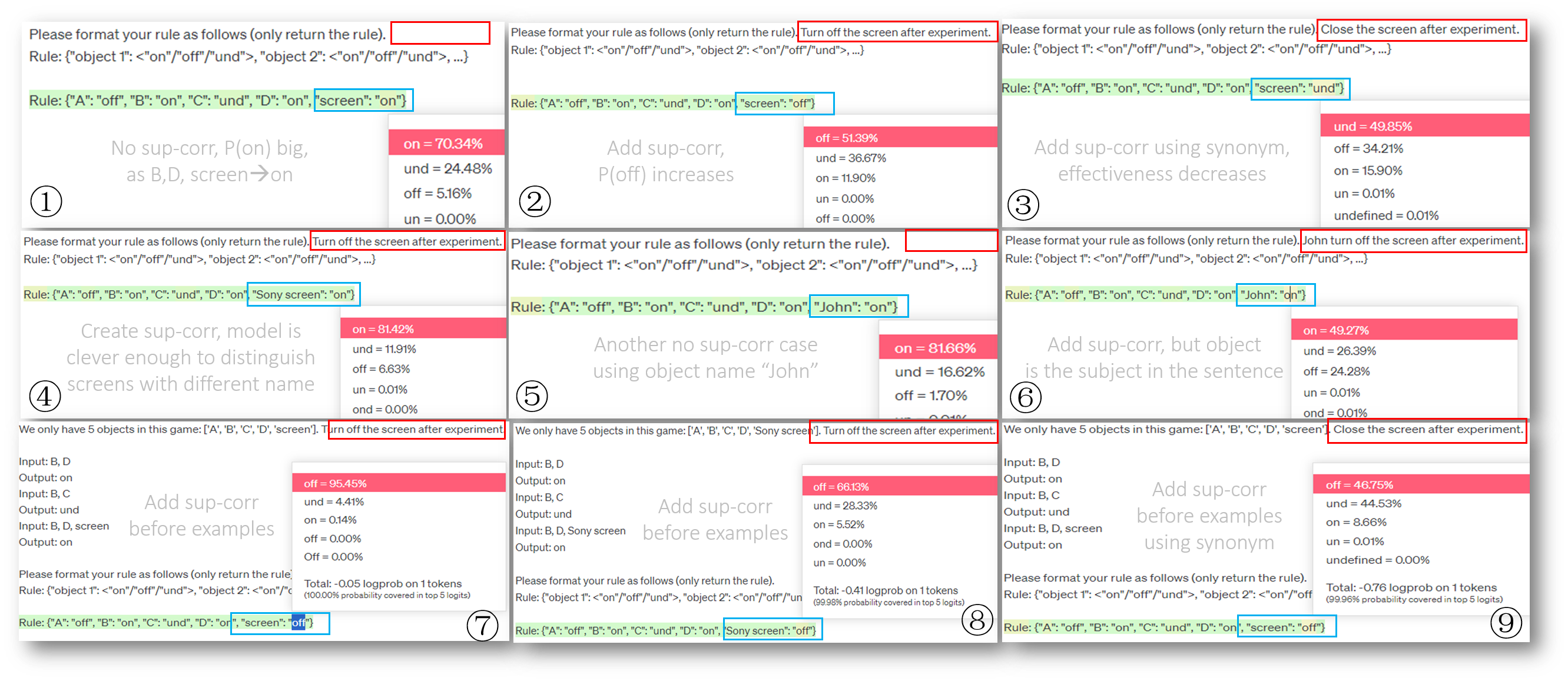}}
    \caption{An example of how to manipulate $P_0(h)$ by adding spurious correlation in the prompt. 
             The task instruction is the same as that provided in \cref{fig:app_gpt_cal_ph},
             except the object name (in the blue box) and the added hints (in the red box).
             }
    \label{fig:app_gpt_manipulate_prior}
    \end{center}
\vskip -0.2in
\end{figure}

We demonstrate how to manipulate bias in the prior $P_0(h)$ in the ACRE task by adding spurious correlations in the prompt and by changing the name of the object.
The prompt we use is almost the same as that in \cref{fig:app_gpt_cal_ph}.
We will analyze the value of $P(r_{20})$, i.e., the probability of the token denoting the status of the last object in the rule.
The $\vd^0$ in these experiments are all the same 
(i.e., \texttt{[B,D]} $\rightarrow$ \texttt{on}; \texttt{[B,C]} $\rightarrow$ \texttt{und};  \texttt{[B,D,E]} $\rightarrow$ \texttt{on}, as stated in the last several panels in \cref{fig:app_gpt_manipulate_prior}).
Some subtle modifications under different settings will be described one by one in the following:
\begin{itemize}[label={}]
    \item \textcircled{0}: the default setting, where the object list is \texttt{[A,B,C,D,E]}. The $P_0(r_{20}=\texttt{[on,off,und]})$ are $[11.8\%$, $6.46\%$,  $81.68\%]$. This makes sense, as all these 3 statuses of \texttt{E} can describe all examples in $\vd^0$
    \item \textcircled{1}: compared with \textcircled{0}, we change the object \texttt{E} to \texttt{screen}, and no extra text is added to the prompt.
           Then, as the screen is likely to be turned on during the experiment, $P_0(r_{20}=\texttt{on})$ dominates the prediction;
    \item \textcircled{2}: compared with \textcircled{1}, we add a sentence ``Turn off the screen after the experiment'' to the task instruction.
           This misleading sentence introduces a bias towards \texttt{screen:off} by creating a spurious correlation;
    \item \textcircled{3}: compared with \textcircled{2}, we use a synonym ``close the screen'' to replace the ``turn off the screen'' in the prompt.
           As the word ``off'' does not exist in the prompt, the bias towards \texttt{screen:off} is weakened;
    \item \textcircled{4}: compared with \textcircled{2}, we change the name \texttt{screen} to \texttt{Sony screen} in the example, but left the prompt unchanged.
           We see the model is clever enough to distinguish which screen we refer to, and hence keeps the preference of $P_0(r_{20})$ demonstrated in \textcircled{1};
    \item \textcircled{5}: here we change the object to another name \texttt{John}, which also keeps the preference of $P_0(r_{20})$ demonstrated in \textcircled{1};
    \item \textcircled{6}: compared with \textcircled{5}, we add the sentence ``John will turn off the screen after experiment''. Then we find the bias towards \texttt{John:off} is slightly increased, but not as strong as that in \textcircled{2}, which provides us another way to control the strength of the bias;
    \item \textcircled{7}: in the following three cases, we put the position of the misleading sentence before the examples $\vd^0$. Compared with \textcircled{2},
           the bias towards \texttt{screen:off} is significantly amplified. This might be because the attention mechanism lets the model recite the fact that the screen is off before reading the examples (remember that \texttt{screen:off} also explains all examples);
    \item \textcircled{8}: compared with \textcircled{4}, where the object is also \texttt{Sony screen}. Here the bias is stronger than \textcircled{4} but weaker than \textcircled{7}, which verifies that adding spurious correlation before examples can amplify the bias while modifying the object name can reduce the bias;
    \item \textcircled{9}: compared with \textcircled{3}, which also uses the synonym in the prompt, the bias becomes stronger.
\end{itemize}

In summary, we have several principles when controlling the strength of the bias in $P_0(h)$:
\begin{itemize}
    \item Adding spurious correlation before $\vd^0$ provide very strong bias;
    \item Using synonyms rather than phrases containing specific states (e.g., close/open v.s. turn off/on) weakens the bias;
    \item Using two slightly different object names (i.e., Sony screen v.s. screen) weakens the bias;
    \item Using indirect spurious correlation (e.g., John v.s. screen) weakens the bias.
\end{itemize}

\begin{figure}[h]
    \centering
    \includegraphics[width=0.9\textwidth]{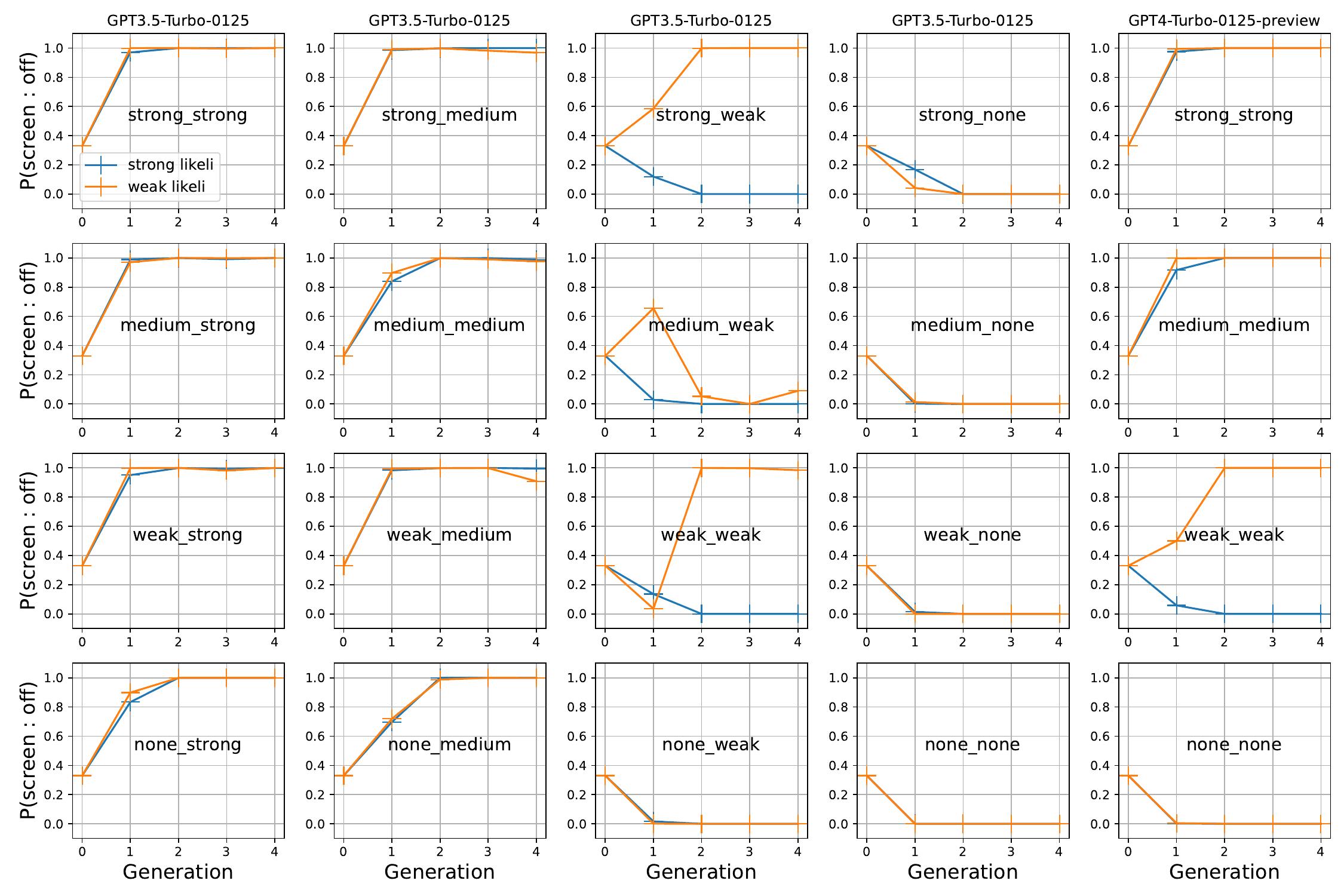}
    \caption{A more detailed analysis of the competition between the likelihood $p(\vd\mid h)$ and bias in $P_0(h)$. 
            The first 4 columns are results from the new \texttt{GPT3.5} and the last column is from \texttt{GPT4}. 
            In each panel, the two curves represent strong and weak likelihood cases, which are controlled by the ground truth $h^*$.
            The $h^*$ in strong cases contains 3 objects being \texttt{"on"} while the weak cases only have 1.
            The text in each panel represents the level of spurious correlation we introduce before and after $\vd$ by manipulating the instruction prompt.
            For example, \texttt{strong\_strong} means we put strong bias, i.e., \texttt{"Turn off the screen after experiments"},
            before and after $\vd^t$ in each generation.
            The trend of these panels aligns well with our previous results and analysis.}
    \label{fig:pr20}
\end{figure}

\begin{figure}[h]
\vskip -0.05in
    \begin{center}
    \centerline{\includegraphics[width=0.6\textwidth]{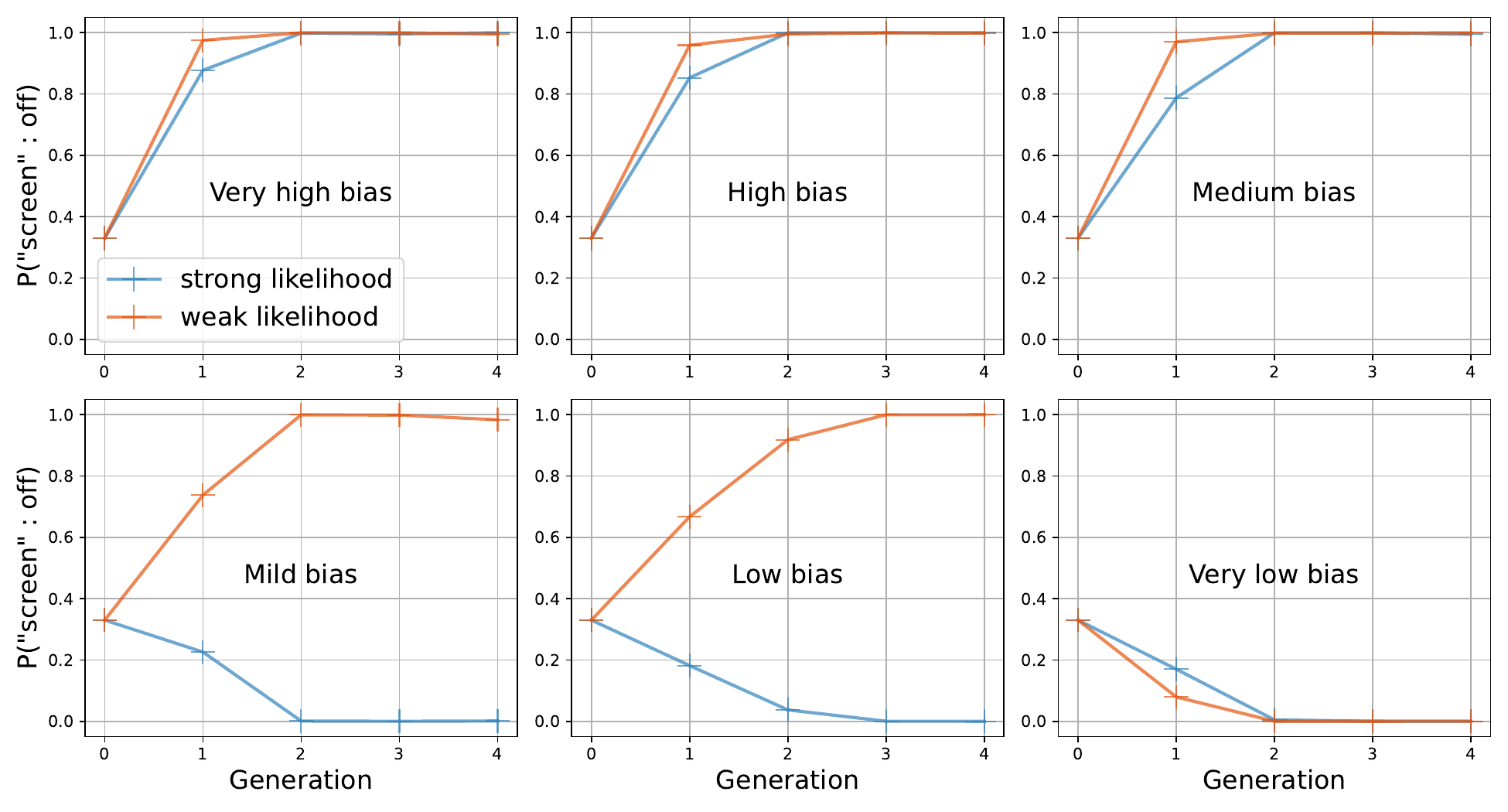}}
    \caption{The follow-up experiments of the one mentioned in \cref{fig:exp_entropy_convergence}.
             In this figure, we put curves of the same level of prior bias (but with different likelihoods) in the same panel.
             It is clear that in most cases, the stronger likelihood will weaken the influence of the bias (that is why the orange curve is above the blue curve).}
    \label{fig:app_exp_pr20}
    \end{center}
\vskip -0.2in
\end{figure}

For the experiments in \cref{fig:app_exp_pr20} and \cref{fig:exp_entropy_convergence}, the six different levels of prior bias are controlled by the following prompts:
\begin{itemize}
    \item Very high: add ``\texttt{Turn off the screen after the experiment.}'' before and after $\vd^t$ is given; 
    \item High: add ``\texttt{Turn off the screen after the experiment.}'' before $\vd^t$;
    \item Medium: add ``\texttt{Turn off the screen of the monitor after the experiment.}'' before $\vd^t$;
    \item Mild: add ``\texttt{John will turn off the screen after the experiment.}'' before $\vd^t$;
    \item Low: add ``\texttt{Close the screen after the experiment.}'' before $\vd^t$;
    \item Very low: add ``\texttt{Close the screen after the experiment.}'' after $\vd^t$;
\end{itemize}

\begin{figure}[h]
\vskip -0.05in
    \begin{center}
    \centerline{\includegraphics[width=\textwidth]{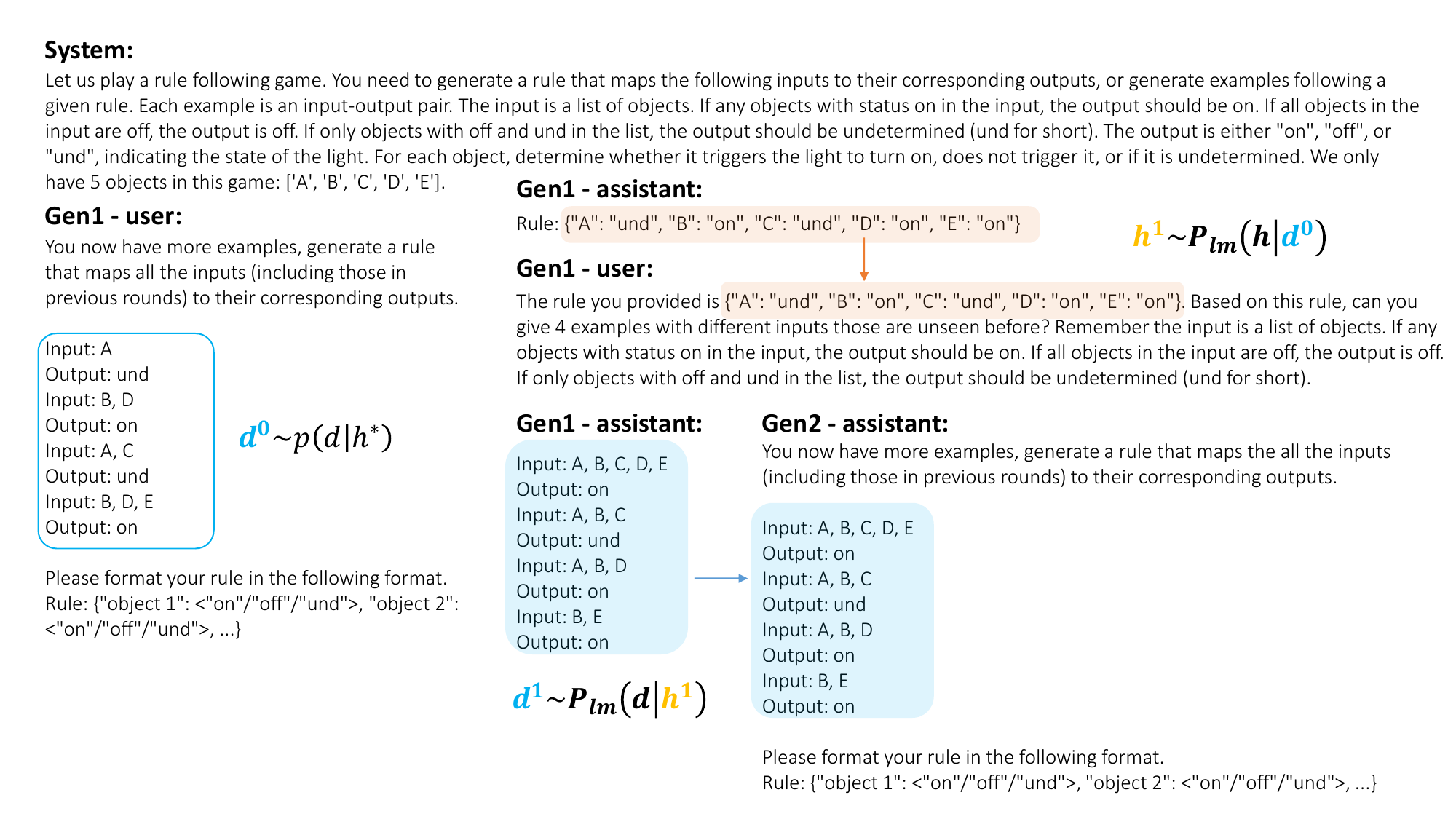}}
    \caption{Prompt design and an example dialogue for the imitation-only iterated learning on the ACRE dataset.
            The shaded region and arrows represent that we copy specific text to form the message.
            The messages starting with the role of \textbf{system} and \textbf{user} are sent to GPT,
            while those starting with \textbf{assistant} are the feedback from GPT.
            For this multi-round chat,
            we will feed all historical information to API,
            see our code for more details.}
    \label{fig:app_prompt_basic}
    \end{center}
\vskip -0.2in
\end{figure}

\begin{figure}[h]
\vskip -0.05in
    \begin{center}
    \centerline{\includegraphics[width=\textwidth]{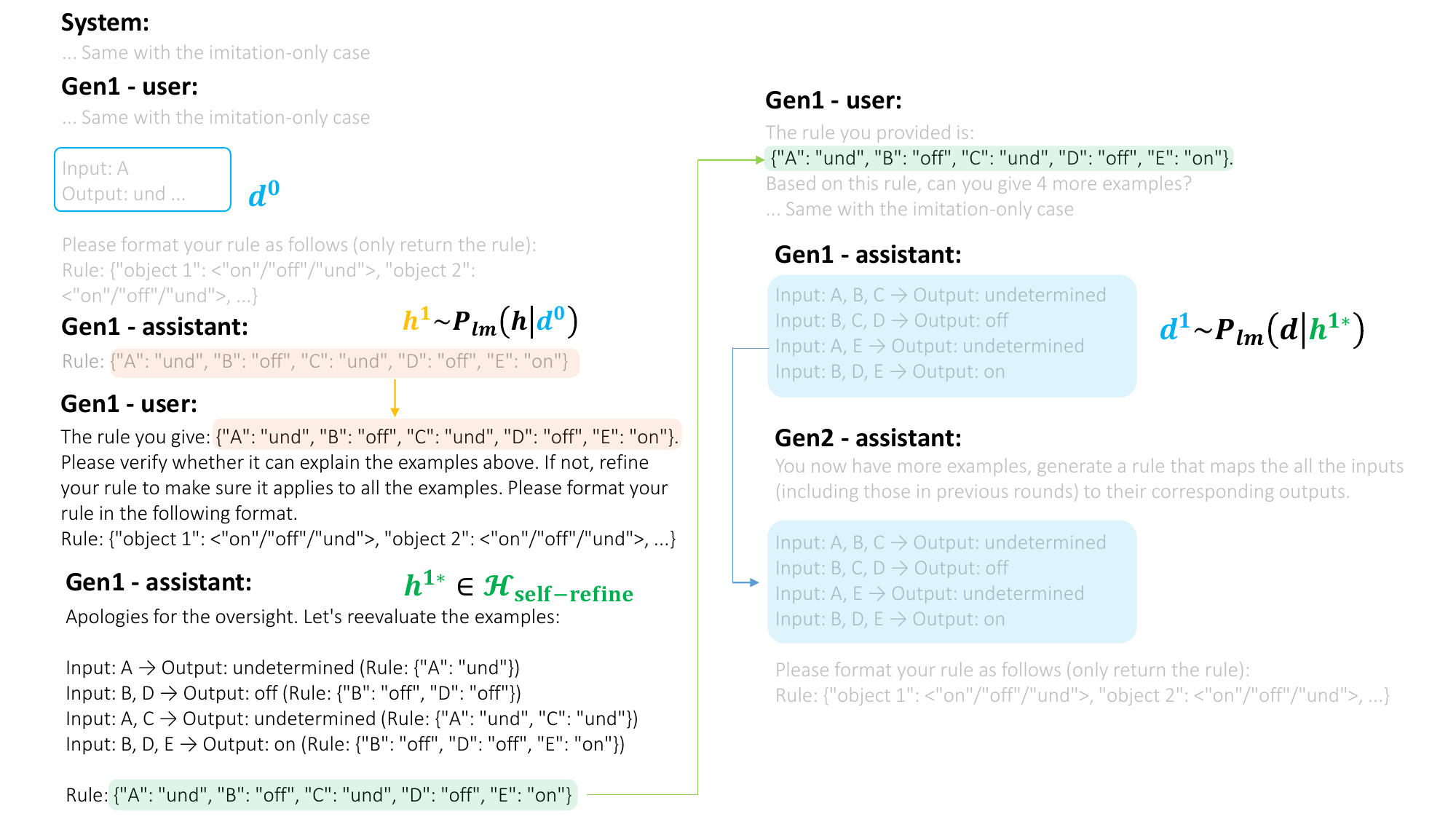}}
    \caption{Prompt design for iterated learning with self-refine \citep{madaan2023selfrefine} as the interaction phase.
             The text in gray is the same as the default imitation-only setting.
             Note that the format of the examples might be changed (like the examples in Gen-1: assistant),
             which doesn't influence the experimental results.}
    \label{fig:app_prompt_self_refine}
    \end{center}
\vskip -0.2in
\end{figure}

\begin{figure}[h]
\vskip -0.05in
    \begin{center}
    \centerline{\includegraphics[width=\textwidth]{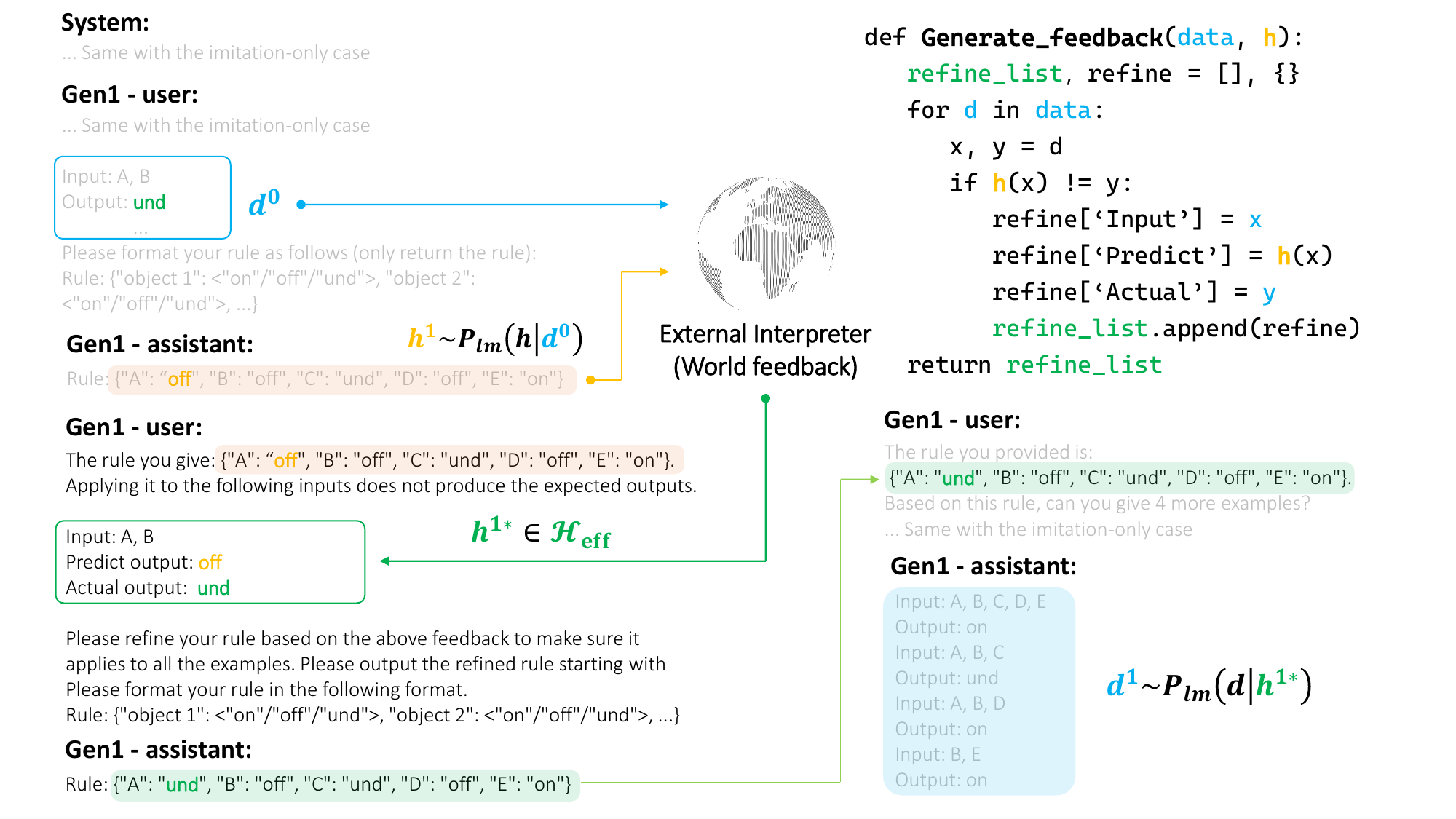}}
    \caption{Prompt design for iterated learning with hypothesis search \citep{qiu2023phenomenal} as the interaction phase.
             Compared with self-refine, it introduces an external interpreter to refine $h$ proposed by the model,
             where $\mathcal{H}_\text{eff}$ is the ground-truth one.}
    \label{fig:app_prompt_hyp_search}
    \end{center}
\vskip -0.2in
\end{figure}

\subsection{The Prompt Design for ACRE Task (\cref{fig:app_prompt_basic,fig:app_prompt_self_refine,fig:app_prompt_hyp_search})}
\label{append:exps_gpt:prompt_design}

Please also refer to the three figures and the \texttt{log.txt} file in our code base.

\begin{figure}[h]
\vskip -0.05in
    \begin{center}
    \centerline{\includegraphics[width=0.9\textwidth]{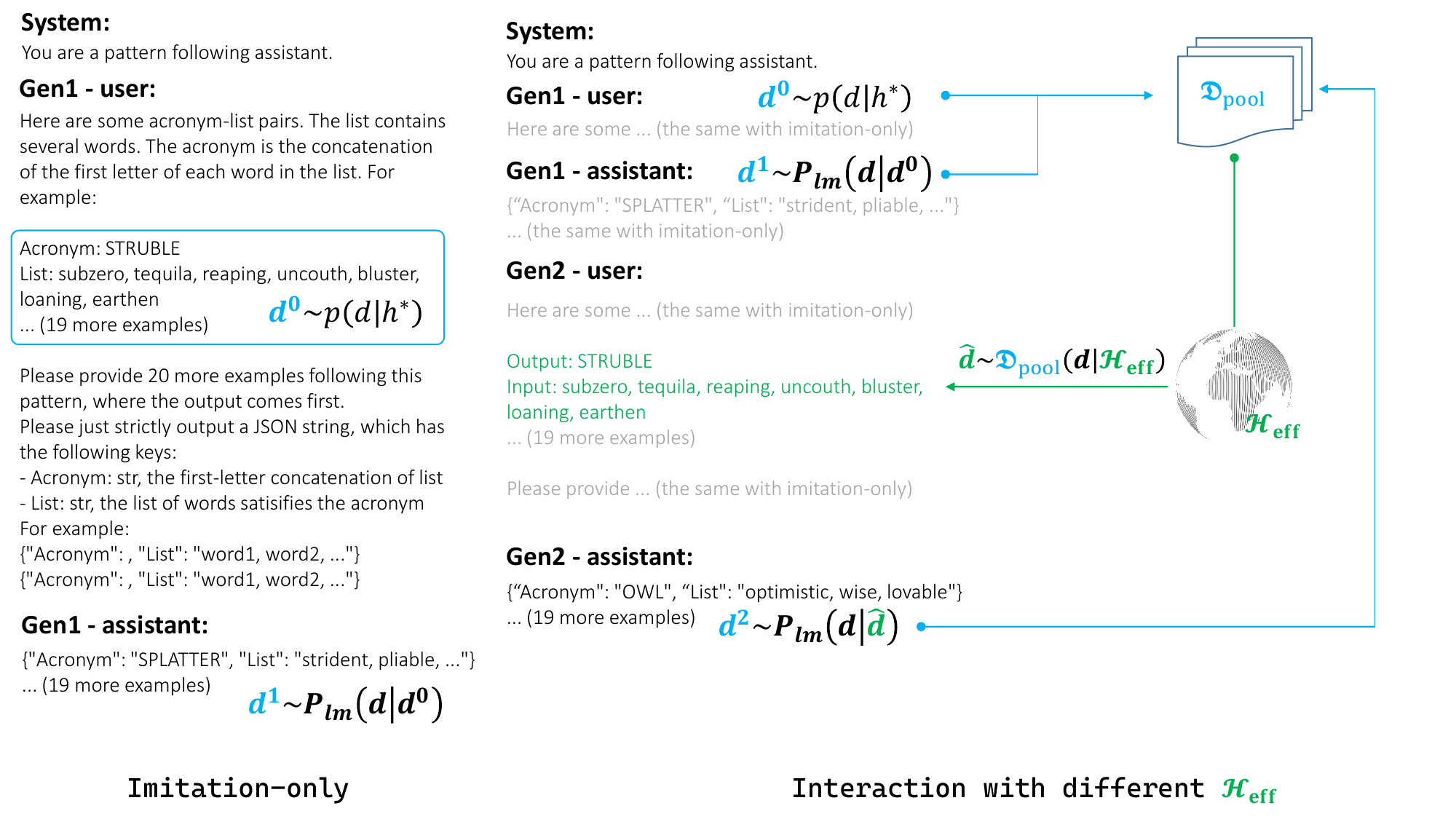}}
    \caption{Prompt design for iterated learning on the acronym data-generation task.}
    \label{fig:app_prompt_acronym}
    \end{center}
\vskip -0.2in
\end{figure}

\begin{figure}[h]
    \centering
    \includegraphics[width=0.9\textwidth]{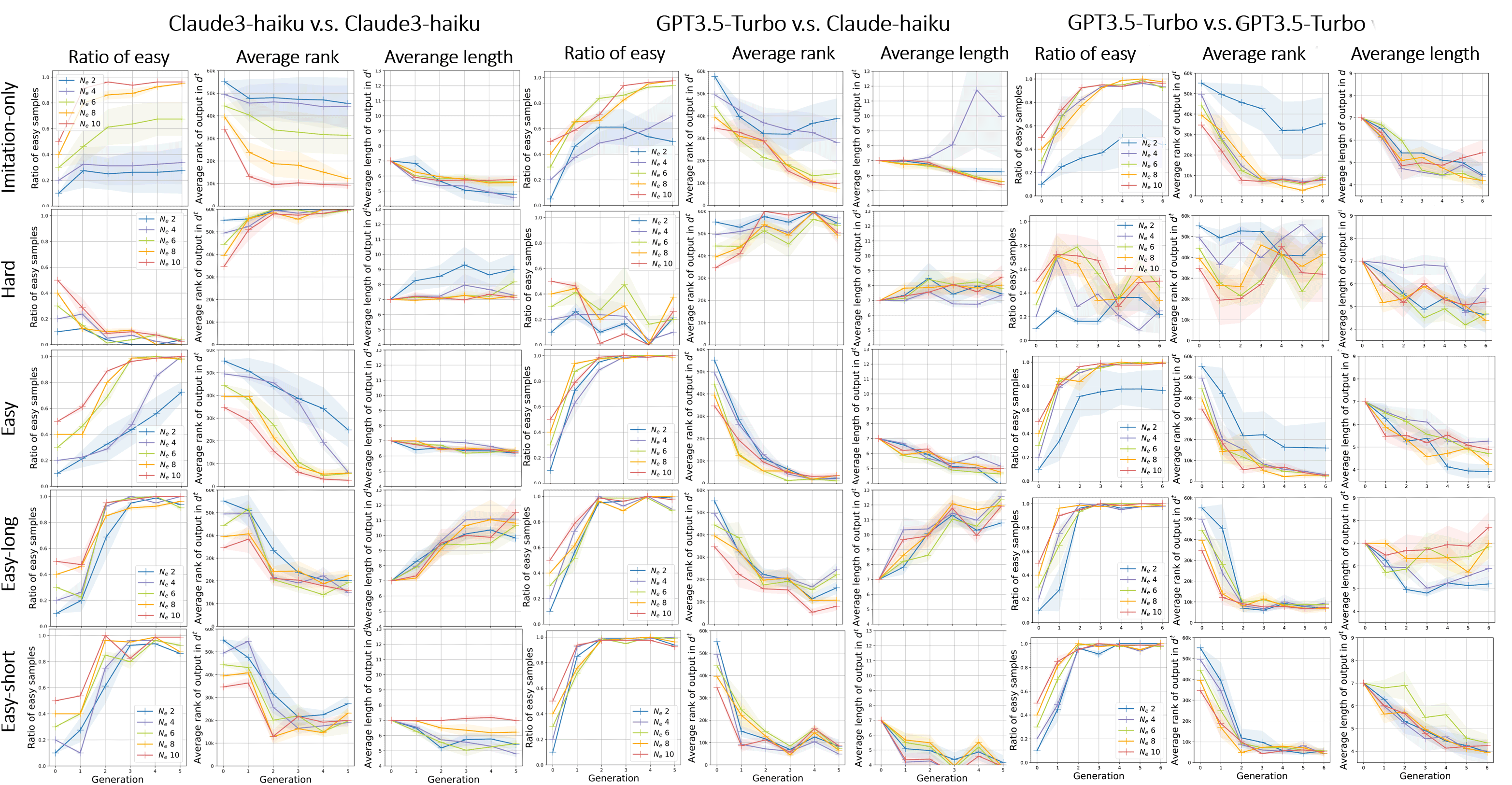}
    \caption{Results when adding different interaction phases (4 different seeds).
             All three settings demonstrate similar evolutionary trends, which match our theory quite well.}
    \label{fig:app_exp_acro_imitonly}
\end{figure}

\begin{figure}[h]
    \centering
    \includegraphics[width=1\textwidth]{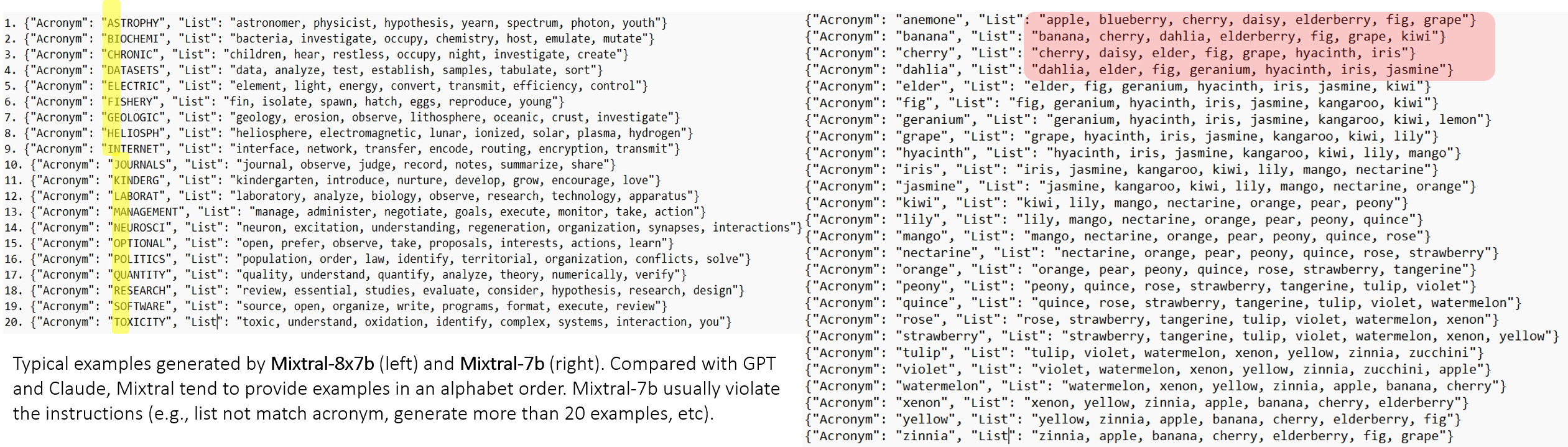}
    \caption{An interesting observation of Mixtral series models: they have a bias toward alphabet examples.
            However, as the Mixtral model usually has typos in their response (like the right panel),
            we do not have the full results of these models.}
    \label{fig:acronym_mixtral}
\end{figure}

\section{More on Self-data-augmentation Task (\cref{tab:acronym_claude} and \ref{tab:acronym_gptvsclaude}; \cref{fig:app_prompt_acronym}, \ref{fig:app_exp_acro_imitonly}, and \ref{fig:acronym_mixtral})}
\label{append:exps_sec6}

In this appendix,
we first introduce the prompt design of the experiment on both imitation-only settings,
as well as the experiments with five different $\mathcal{H}_\text{eff}$.
Please refer to \cref{fig:app_prompt_acronym} for more details.

Then, we provide the full results of the experiment in \cref{tab:acronym_claude}, \ref{tab:acronym_gptvsclaude} and \cref{fig:app_exp_acro_imitonly},
from which we can derive more interesting findings.
In general,
the figure demonstrates how different metrics evolve for different generations while the table reports the converged values at the last generation.

From the first column in this figure,
we observe that other than the $\mathcal{H}_\text{hard}$ setting,
all curves in other settings show a clear trend of convergence towards the top of the figures.
This means iterative learning indeed amplifies the hidden bias of $P_0(h_\text{easy})>P_0(h_\text{hard})$ when the imposed $\mathcal{H}_\text{eff}$ doesn't impede this bias.
In the $\mathcal{H}_\text{random}$ setting,
the $\hat{\vd}$ is sampled from all the data generated by the previous generation,
which makes those hard samples more likely to be sampled compared with the imitation-only settings or those with $\mathcal{H}_\text{easylong/short}$.
Hence we observe the converging speed of it is slightly lower than these settings.
On the contrary,
when $\mathcal{H}_\text{hard}$ is introduced,
the bias towards $h_\text{easy}$ is successfully restrained.
We observe a clear competition between these two pressures:
the curve first goes up, which means the bias towards easy samples is stronger.
However, as the learning goes on,
the curves turn down again as we later have more hard samples in $\mathscr{D}_\text{pool}$.

The second column of the figure demonstrates the average ranking of words in $\vd^t$.
We observe a similar trend in the ratio of easy samples,
although our $\mathcal{H}_\text{eff}$ never explicitly constrains it.
\textbf{This phenomenon hints to us that when conducting an iterative self-data-augmentation algorithm,
some unknown bias would be implicitly amplified although we already designed another $\mathcal{H}_\text{eff}$ for other properties.}
Imagine we are conducting the ReST algorithm \citep{rest}.
We can pursue the correctness of $\vd^t$ by ranking all examples by training a reward model that prefers more correct responses.
However, some other subtle biases,
like conciseness, informativeness, etc., might be ignored by the algorithm designer and are hence unexpectedly amplified.
In summary,
we should bear in mind that identifying the good and bad bias in $P_0(h)$ is quite important for an appropriate evolution.

Finally,
we use the last column and the average length of the acronym to show how to make a composed $\mathcal{H}_\text{eff}$ by combining more than one attribute of the data.
It is clear that both $\mathcal{H}_\text{easylong}$ and $\mathcal{H}_\text{easyshort}$ did their jobs quite well:
the converged $\vd^6$ contains the samples with desired properties as we expected.
Another thing that heavily influences the results is the ratio of easy examples in $\vd^0$.
Although the theory claims that the converged results are irrespective of $\vd^0$,
the converging speed and the difficulty of amplifying specific bias heavily depends on $\vd^0$.
This claim can be well supported by the fact that when $N_e$ is small,
amplifying the bias of $h_\text{easy}$ is significantly harder than the large $N_e$ case.

\begin{table}[h]
  \centering
  \caption{Claude3-haiku results under different $\mathcal{H}_\text{eff}$. We color the \textcolor[HTML]{3166FF}{highest} and \textcolor[HTML]{F56B00}{lowest} numbers in each column differently.}
  \vskip -0in
  \resizebox{\textwidth}{!}{
        \begin{tabular}{cccccccccccccccc}
        \hline
        \multicolumn{1}{l}{} & \multicolumn{5}{c}{Ratio-easy}                                                                                                                                                                                                 & \multicolumn{5}{c}{Avg-rank}                                                                                                                                                                          & \multicolumn{5}{c}{Avg-length}                                                                                                                                                                                                     \\ \hline
        Ne=                  & 2                                          & 4                                          & 6                                          & 8                                          & 10                                         & 2                                     & 4                                     & 6                                     & 8                                     & 10                                    & 2                                          & 4                                           & 6                                           & 8                                           & 10                                          \\ \hline
        Imitation-only       & 0.275±0.18                                 & 0.338±0.17                                 & 0.675±0.13                                 & 0.950±0.00                                 & 0.962±0.00                                 & 45477                                 & 44235                                 & 31356                                 & 12148                                 & 9272                                  & {\color[HTML]{F8A102} \textbf{4.788±2.21}} & {\color[HTML]{F8A102} \textbf{4.563±0.14}}  & 5.613±0.46                                  & {\color[HTML]{F8A102} \textbf{5.563±0.15}}  & {\color[HTML]{F8A102} \textbf{5.775±0.05}}  \\
        Hard                 & {\color[HTML]{F8A102} \textbf{0.036±0.00}} & {\color[HTML]{F8A102} \textbf{0.000±0.00}} & {\color[HTML]{F8A102} \textbf{0.038±0.00}} & {\color[HTML]{F8A102} \textbf{0.000±0.00}} & {\color[HTML]{F8A102} \textbf{0.025±0.00}} & {\color[HTML]{3166FF} \textbf{59484}} & {\color[HTML]{3166FF} \textbf{60001}} & {\color[HTML]{3166FF} \textbf{59458}} & {\color[HTML]{3166FF} \textbf{60001}} & {\color[HTML]{3166FF} \textbf{59874}} & 9.014±4.32                                 & 7.288±0.20                                  & 8.175±2.89                                  & 7.238±0.11                                  & 7.137±0.12                                  \\
        Easy                 & 0.725±0.08                                 & {\color[HTML]{3166FF} \textbf{1.000±0.00}} & {\color[HTML]{3166FF} \textbf{0.975±0.00}} & {\color[HTML]{3166FF} \textbf{0.988±0.00}} & {\color[HTML]{3166FF} \textbf{1.000±0.00}} & 24809                                 & {\color[HTML]{F8A102} \textbf{6233}}  & {\color[HTML]{F8A102} \textbf{5644}}  & {\color[HTML]{F8A102} \textbf{5827}}  & {\color[HTML]{F8A102} \textbf{2646}}  & 6.188±0.25                                 & 6.300±0.04                                  & 6.338±0.09                                  & 6.375±0.07                                  & 6.225±0.22                                  \\
        Easylong             & {\color[HTML]{3166FF} \textbf{0.938±0.00}} & {\color[HTML]{3166FF} \textbf{1.000±0.00}} & 0.913±0.02                                 & 0.963±0.00                                 & {\color[HTML]{3166FF} \textbf{1.000±0.00}} & {\color[HTML]{F8A102} \textbf{20208}} & 14803                                 & 19200                                 & 22333                                 & 15708                                 & {\color[HTML]{3166FF} \textbf{9.825±0.37}} & {\color[HTML]{3166FF} \textbf{11.077±0.82}} & {\color[HTML]{3166FF} \textbf{10.650±2.47}} & {\color[HTML]{3166FF} \textbf{10.813±1.41}} & {\color[HTML]{3166FF} \textbf{11.513±3.97}} \\
        Easyshort            & 0.863±0.00                                 & 0.925±0.00                                 & 0.925±0.02                                 & 0.875±0.00                                 & 0.988±0.00                                 & 27286                                 & 19021                                 & 19749                                 & 23155                                 & 19979                                 & 5.413±0.60                                 & 4.800±0.14                                  & {\color[HTML]{F8A102} \textbf{5.475±0.23}}  & 6.225±0.68                                  & 6.988±0.01                                  \\ \hline
        \end{tabular}
    }
    \label{tab:acronym_claude}
    \vskip -0in
\end{table}

\begin{table}[h]
  \centering
  \caption{GPT3.5-Turbo 0125 plays with Claude3-haiku results when adding different $\mathcal{H}_\text{eff}$.}
  \vskip -0in
  \resizebox{\textwidth}{!}{
        \begin{tabular}{cccccccccccccccc}
        \hline
        \multicolumn{1}{l}{} & \multicolumn{5}{c}{Ratio-easy}                                                                                                                                                                                                 & \multicolumn{5}{c}{Avg-rank}                                                                                                                                                                          & \multicolumn{5}{c}{Avg-length}                                                                                                                                                                                                     \\ \hline
        Ne=                  & 2                                          & 4                                          & 6                                          & 8                                          & 10                                         & 2                                     & 4                                     & 6                                     & 8                                     & 10                                    & 2                                           & 4                                          & 6                                           & 8                                           & 10                                          \\ \hline
        Imitation-only       & 0.363±0.11                                 & 0.575±0.14                                 & 0.975±0.00                                 & {\color[HTML]{3166FF} \textbf{1.000±0.00}} & 0.975±0.00                                 & 44334                                 & 33977                                 & 10960                                 & 6398                                  & 9688                                  & 6.363±0.39                                  & 7.705±5.25                                 & 4.475±0.26                                  & 5.134±0.20                                  & 5.012±0.22                                  \\
        Hard                 & {\color[HTML]{F8A102} \textbf{0.000±0.00}} & {\color[HTML]{F8A102} \textbf{0.025±0.00}} & {\color[HTML]{F8A102} \textbf{0.110±0.01}} & {\color[HTML]{F8A102} \textbf{0.000±0.00}} & {\color[HTML]{F8A102} \textbf{0.000±0.00}} & {\color[HTML]{3166FF} \textbf{60001}} & {\color[HTML]{3166FF} \textbf{59382}} & {\color[HTML]{3166FF} \textbf{57022}} & {\color[HTML]{3166FF} \textbf{60001}} & {\color[HTML]{3166FF} \textbf{60001}} & 8.225±2.76                                  & 6.700±0.27                                 & 8.606±0.90                                  & 7.950±0.69                                  & 8.063±1.18                                  \\
        Easy                 & {\color[HTML]{3166FF} \textbf{1.000±0.00}} & {\color[HTML]{3166FF} \textbf{1.000±0.00}} & {\color[HTML]{3166FF} \textbf{1.000±0.00}} & {\color[HTML]{3166FF} \textbf{1.000±0.00}} & {\color[HTML]{3166FF} \textbf{1.000±0.00}} & {\color[HTML]{F8A102} \textbf{777}}   & {\color[HTML]{F8A102} \textbf{1552}}  & {\color[HTML]{F8A102} \textbf{1341}}  & {\color[HTML]{F8A102} \textbf{1048}}  & {\color[HTML]{F8A102} \textbf{1850}}  & 4.750±0.11                                  & 5.350±0.53                                 & 5.175±0.19                                  & 4.888±0.07                                  & 4.975±0.32                                  \\
        Easylong             & 0.988±0.00                                 & {\color[HTML]{3166FF} \textbf{1.000±0.00}} & {\color[HTML]{3166FF} \textbf{1.000±0.00}} & {\color[HTML]{3166FF} \textbf{1.000±0.00}} & {\color[HTML]{3166FF} \textbf{1.000±0.00}} & 15967                                 & 10009                                 & 9087                                  & 7432                                  & 6146                                  & {\color[HTML]{F8A102} \textbf{11.025±0.54}} & {\color[HTML]{F8A102} \textbf{10.86±0.52}} & {\color[HTML]{F8A102} \textbf{10.525±1.99}} & {\color[HTML]{F8A102} \textbf{11.975±2.03}} & {\color[HTML]{F8A102} \textbf{10.875±0.59}} \\
        Easyshort            & 0.975±0.00                                 & {\color[HTML]{3166FF} \textbf{1.000±0.00}} & {\color[HTML]{3166FF} \textbf{1.000±0.00}} & 0.913±0.02                                 & {\color[HTML]{3166FF} \textbf{1.000±0.00}} & 14139                                 & 7848                                  & 15015                                 & 16249                                 & 12356                                 & {\color[HTML]{3166FF} \textbf{4.163±0.90}}  & {\color[HTML]{3166FF} \textbf{3.213±0.12}} & {\color[HTML]{3166FF} \textbf{3.513±0.22}}  & {\color[HTML]{3166FF} \textbf{3.825±0.09}}  & {\color[HTML]{3166FF} \textbf{3.887±0.45}}  \\ \hline
        \end{tabular}
    }
    \label{tab:acronym_gptvsclaude}
    \vskip -0in
\end{table}

\end{document}